\documentclass[Afour,sageh,times]{sagej}
\usepackage{moreverb, url}

\usepackage{basicmath}
\usepackage{graphicx}   
\usepackage{amsmath} 
\usepackage{amssymb}  
\usepackage{units}
\usepackage[normalem]{ulem}

\usepackage{subcaption}

\usepackage{color} 

\usepackage{enumitem}
\usepackage{booktabs}

\usepackage[
colorlinks,
bookmarksopen,
bookmarksnumbered,
citecolor=red,
urlcolor=red
]{hyperref}

\usepackage[capitalize]{cleveref}

\newcommand\BibTeX{{\rmfamily B\kern-.05em \textsc{i\kern-.025em b}\kern-.08em
		T\kern-.1667em\lower.7ex\hbox{E}\kern-.125emX}}

\newcommand{\change}[1]{\textcolor{blue}{#1}}
\renewcommand{\change}[1]{#1}
\title{Impact-Invariant Control: Maximizing Control Authority During Impacts}

\author{William Yang and Michael Posa}

\newcommand{\genPos}{q}
\newcommand{\genPosPre}{q^-}
\newcommand{\genPosPost}{q^+}
\newcommand{\genVel}{v}
\newcommand{\genVelPre}{v^-}
\newcommand{\genVelPost}{v^+}

\newcommand{\genAcc}{\dot{v}}
\newcommand{\genState}{x}
\newcommand{\genStatePre}{x^-}
\newcommand{\genStatePost}{x^+}
\newcommand{\taskSpacePos}{y}
\newcommand{\taskSpacePosDes}{y_{des}}
\newcommand{\taskSpaceVel}{\dot{y}}
\newcommand{\taskSpaceVelDes}{\dot{y}_{des}}
\newcommand{\taskSpaceAcc}{\ddot{y}}
\newcommand{\taskSpaceAccDes}{\ddot{y}_{des}}

\newcommand{\contactJacobian}{J_{\lambda}}
\newcommand{\taskSpaceJacobian}{J_{y}}

\newcommand{\holonomicJacobian}{J_{h}}
\newcommand{\contactImpulse}{\Lambda}
\newcommand{\nominalControl}{u}
\newcommand{\feedforwardControlTerm}{u_{ff}}
\newcommand{\feedbackControlTerm}{u_{fb}}

\newcommand{\projVel}{v_{proj}}
\newcommand{\projTaskSpaceVel}{\dot{y}_{proj}}
\newcommand{\projTaskSpaceVelError}{\dot{\tilde{y}}_{proj}}

\begin{document}

	\begin{abstract}
		When legged robots impact their environment executing dynamic motions, they undergo large changes in their velocities in a short amount of time.
		Measuring and applying feedback to these velocities is challenging, further complicated by uncertainty in the impact model and impact timing.
		This work proposes a general framework for adapting feedback control during impact by projecting the control objectives to a subspace that is invariant to the impact event.
		The resultant controller is robust to uncertainties in the impact event while maintaining maximum control authority over the impact-invariant subspace.
		We demonstrate the improved performance of the projection over other commonly used heuristics on a walking controller for a planar five-link-biped.
		The projection is also applied to jumping, box jumping, and running controllers for the compliant 3D bipedal robot, Cassie.
		The modification is easily applied to these various controllers and is a critical component to deploying on the physical robot.
		Code and video of the experiments are available at \href{https://impact-invariant-control.github.io/}{https://impact-invariant-control.github.io/}.
	\end{abstract}
	\maketitle
	
	\section{Introduction}
	\label{sec:introduction}
	
	Handling the making and breaking of contact lies at the core of controllers for legged robots.
	Recent advances in the modeling and planning of these contacts have enabled legged robots to walk reliably in select environments \cite{wensing2023optimization}.
	This progress directs the focus of the field toward developing legged robots capable of increasingly agile motions.
	However, these agile motions often require interacting with the environment with non-negligible impact events, something our current controllers are incredibly sensitive to.
	When a robot's foot makes contact with the world, the foot is brought quickly to a stop by a large contact impulse.
	Large contact forces and rapidly changing velocities hinders accurate state estimation.
	Coupled with the relatively poor predictive performance of our contact models \cite{halm2019modeling} \cite{fazeli2020fundamental} \cite{remy2017ambiguous}, this combination of large state uncertainty and poor models makes control especially difficult.
	
	Roboticists have attempted to improve the robustness of legged locomotion to these impact events by addressing the reference trajectories as well as the controllers that track those trajectories.
	For example, the open-loop swing-leg retraction policy has been shown to have inherent stability to varying terrain heights \cite{seyfarth2003swing}. 
	Qualitatively similar motions were also found independently through robust trajectory optimization \cite{dai2012optimizing} \cite{green2020planning} \cite{zhu2022hybrid}.
	While designing more robust trajectories shows promise, the challenge of designing controllers to track these often discontinuous trajectories still remains.
	
	\begin{figure}[t!]
		\centering
		\includegraphics[width=0.48\textwidth]{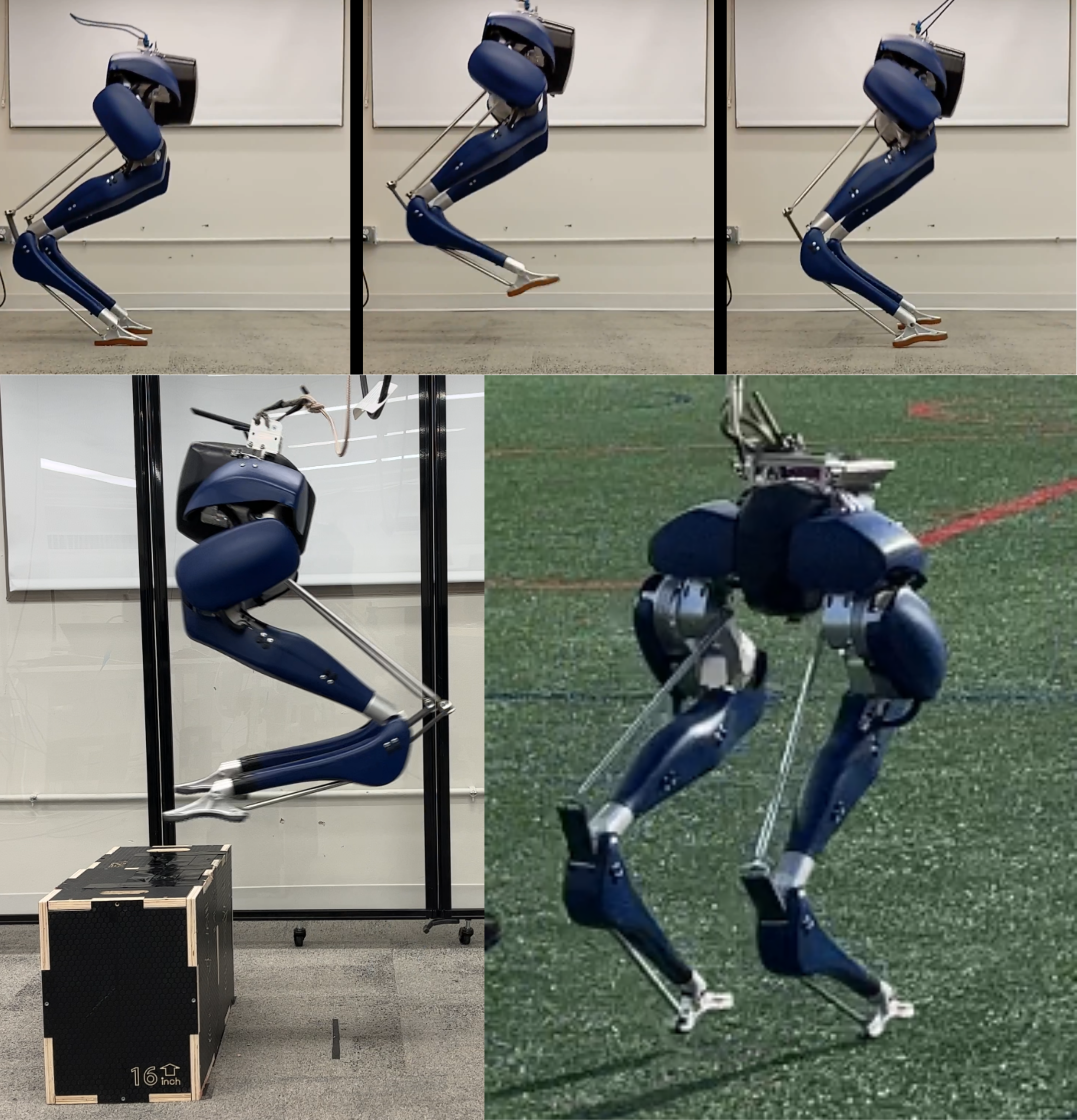}
		\caption{Cassie is able to execute agile motions with non-negligible impacts like jumping (top), box jumping (bottom-left), and running (bottom-right) using impact-invariant control. }
		\label{fig:cassie_hardware_jump}
	\end{figure}
	
	Tracking a discontinuous trajectory is problematic due to the unavoidable difference between the reference trajectory and actual system caused by even minuscule differences in impact timing.
	These differences cause feedback control efforts to spike and optimization constraints to be violated \cite{wang2019impact}, leading to instabilities.
	
	\change{Alternative approaches focus instead on avoiding discontinuities from impact altogether.
		For example, the linear inverted pendulum (LIP) \cite{kajita20013d} is a popular template that does not have center of mass velocity in the vertical direction.
		Another popular simple model is the spring-loaded inverted pendulum (SLIP) \cite{schwind1998spring} has massless feet which does not produce discontinuous trajectories.}
	While impacts do not appear in these simple models, impacts will still manifest when embedding these templates onto physical robots with non-negligible mass in the legs.
	Furthermore, it is neither possible nor desirable to avoid impacts for more agile motions such as running or jumping.
	Thus, handling non-trivial impacts in a robust manner is essential to the development of more agile legged robots.
	
	In this work, we propose a method for tracking discontinuous trajectories across impacts that directly avoids jumps in tracking error.
	We achieve this by projecting the tracking objectives down to a subspace where they are invariant to the impact event.
	\cite{gong2022zero} shared an important insight about angular momentum about the contact point, noting that it is invariant to impacts at that contact point.
	Inspired by this, we generalize this property and extend it to include the entire invariant subspace, which we term the impact-invariant subspace.
	
	\change{
		This work makes the following contributions:
		\begin{itemize}
			\item The primary contribution of this paper is the identification and understanding of a subspace of velocities that are invariant to contact impulses. 
			\item We propose a method that modifies controller feedback that leverages this subspace to improve robustness to impact uncertainty.
			Our method can be applied to general task-space tracking objectives for even complex robots such as the bipedal robot Cassie and can be computed fast enough for kilohertz rate control.
			\item We show that our method can derived from a robust optimal control perspective.
			\item We extensively evaluate our method across a range of controllers for bipedal robots.
			Included in these examples, validated on the physical bipedal Cassie robot, are a 40 cm jump up onto a box and running on a turf field. To the best of our knowledge, this is the first example of a model-based controller for Cassie for both box jumping and running.
		\end{itemize}
		A preliminary version of this article was presented at the International Conference on Intelligent Robots and System 2021 \cite{yang2021impact}.
		The contributions made in this paper extend from the conference paper in two main aspects.
		First, this paper provides a derivation of impact-invariant control from a robust optimal control perspective under reasonable assumptions.
		The derivation shows that the original formulation proposed in the conference paper is complete and does not need to be modified.
		For this reason, the remaining extensions in this paper lie within extensive new examples, which are more dynamic jumping controllers and a running controller for Cassie.
		Both the box jumping controller and running controller introduced in this paper are the first model-based controllers for their respective motions demonstrated on Cassie.
	}

	\section{Related Work}
	\label{sec:related_work}
	
	\change{
		A primary challenge of control during impacts is avoiding erroneous controller spikes that are a result of applying feedback to discontinuous trajectories.
		While controller spikes can be reduced through strategies such as reducing controller gains and contact constraints around the impact event \cite{mason2016balancing} \cite{atkeson2015no}, these heuristic methods do not address the fundamental challenge of tracking discontinuous trajectories.
		A strategy that does attempt to directly address this challenge is termed reference spreading control \cite{rijnen2017control} \cite{van2022robot}.
		This method extends the reference trajectories to ensure that a valid reference trajectory exists despite mismatches in impact timing.
		An extension to reference spreading control \cite{van2023dual} eliminates the time-dependency through vector fields.
		However, all of these methods rely on turning off all velocity feedback while the impact is still resolving.
	}
	
	\change{
		Another impact-aware control formulation is \cite{wang2019impact}, which incorporates anticipated impacts into a robust control formulation. 
		When evaluating velocity constraints near impacts, the authors consider both the pre- and post-impact velocities with an analytical and empirical approach \cite{wang2022predicting} in their quadratic program (QP) controller.
		The primary distinction is that \cite{wang2019impact} addresses how to account for and avoid violation of hardware velocity constraints through an impact event, while our work focuses on how to properly track discontinuous signals across impacts. 
	}
	
	\change{
		\cite{rijnen2017control} and \cite{kong2023saltation} both leverage the saltation matrix to properly perform linearizations across hybrid events.
		While a saltation matrix informs how to propagate the proper linear feedback gains for a hybrid system such as in Hybrid LQR, similar to other works it does not apply to control during the hybrid transition.
	}
	
	\change{
		Control \textit{during} is the key aspect of our work.
		While prior works assumes impacts resolve instantaneously and therefore only consider pre- and post-impacts states, impacts on real systems can take tens of milliseconds to resolve \cite{acosta2022validating}, which is a non-negligible amount of time in the timescale of tracking controllers.
		Our proposed method makes no assumptions on the magnitude or duration of impact forces, which allows our method to handle when the impact is still resolving.
	}
	
	\section{Background}
	\label{sec:background}
	
	We briefly introduce preliminary material and notation from both multibody dynamics and optimization-based control.
	
	\subsection{Robot Models}
	
\begin{figure}
		\includegraphics[width=0.20\textwidth]{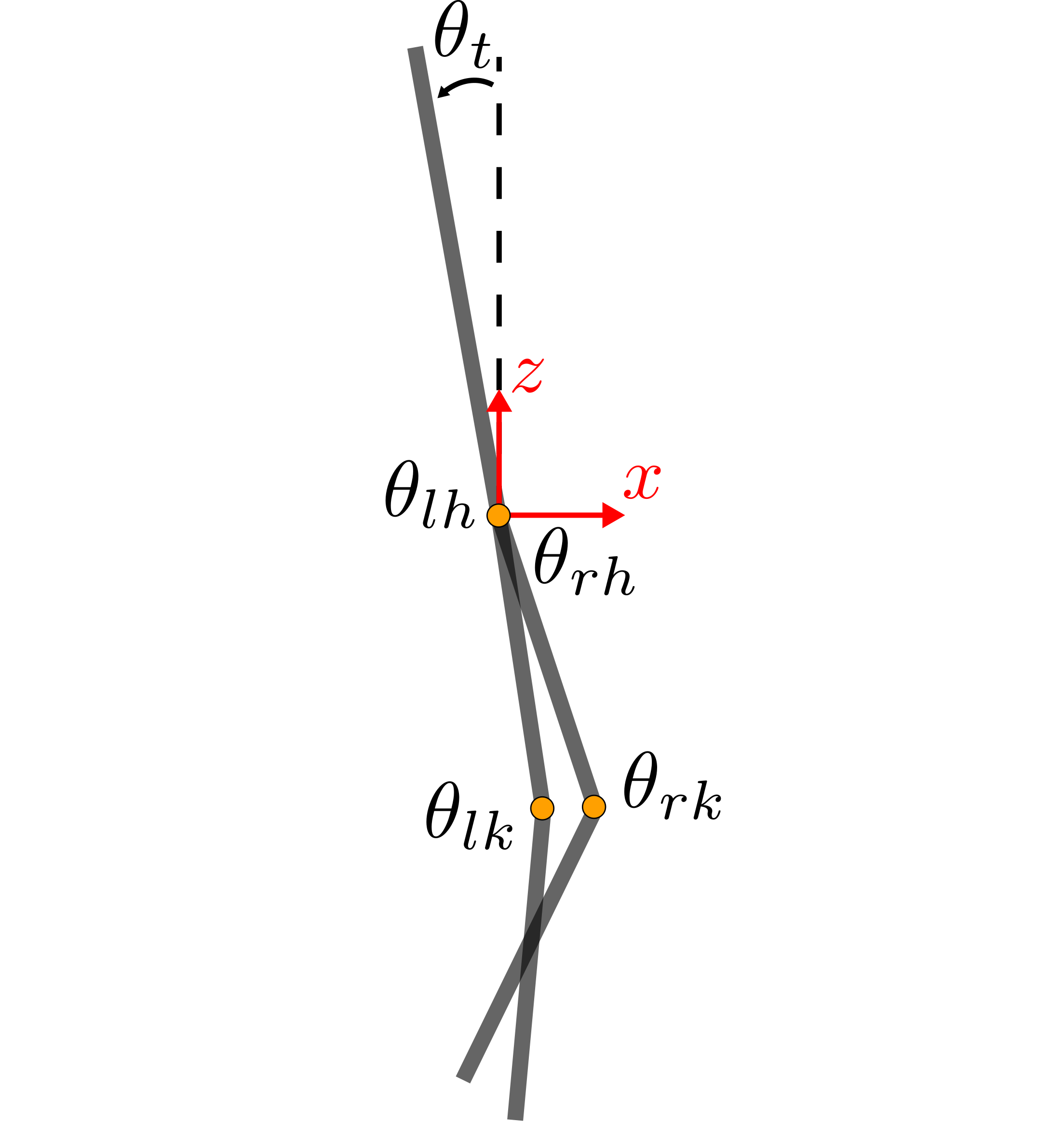}
		\includegraphics[width=0.28\textwidth]{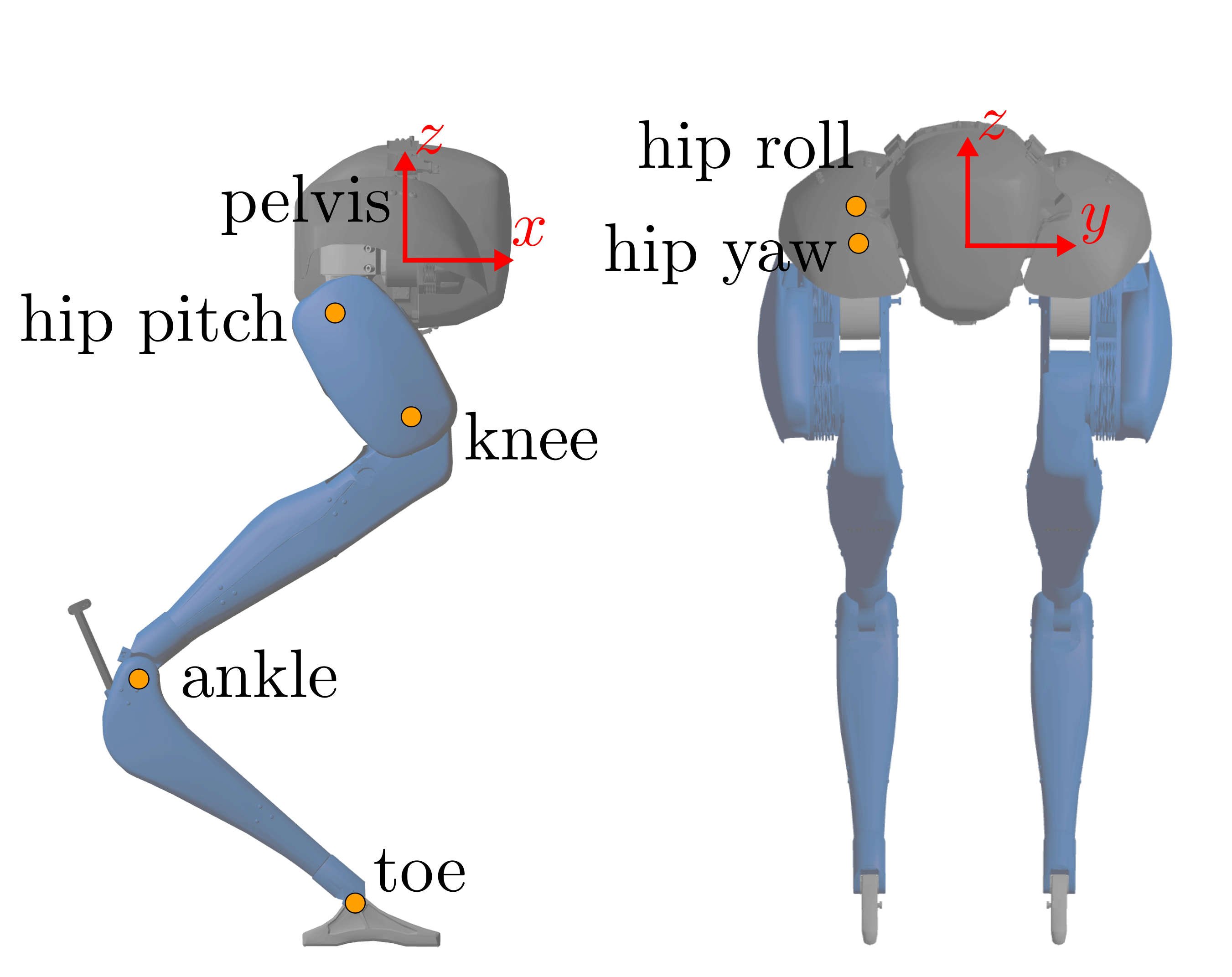}
		\caption{\change{We use both the planar biped Rabbit (left) and the 3D compliant bipedal robot Cassie (right) as concrete examples to highlight the advantages of impact-invariant control.}}
		\label{fig:robot_models}
	\end{figure}

	\change{
		We consider two legged robots, the five-link planar biped Rabbit \cite{chevallereau2003rabbit} and the 3D compliant bipedal robot Cassie shown in \cref{fig:robot_models}, to ground impact-invariant control in concrete applications.
		Both legged robots are modeled using conventional floating-base Lagrangian rigid-body dynamics.
	}
	
	\change{
		Rabbit has 7 degrees of freedom with 4 degrees of actuation while Cassie has 22 degrees of freedom with 10 degrees of actuation.
		Cassie has 4 physical leaf springs located at its ankle and knee joints. 
		In our controllers, we treat these springs as rigid, reducing our controller model to 18 degrees of freedom.
		However when evaluating our results in simulation, we do include the springs, modeling them as torsional springs located directly at the ankle and knee joints.
	}
	
	\subsection{Rigid Body Dynamics}
	
	\change{
		The robot's state $\genState \in \Real^{n_q + n_v} = \left[\genPos; \genVel\right]$, described by its positions $\genPos \in \Real^{n_q}$ and velocities $\genVel \in \Real^{n_v}$, is expressed in generalized floating-base coordinates.
		We use quaternions to represent the 3D orientation of Cassie.
		For this reason, the time derivatives of the positions, $\dot q$, are related to the velocities, $\genVel$ by the map $N(q) \in \Real^{n_q \times n_v}$,
		\begin{align}
			\dot q = N(q) \genVel.
		\end{align}
	}
	For Rabbit, $N(q)$ is the identity matrix, but is not for Cassie.
	The dynamics are derived using the Euler-Langrange equation and expressed in the form of the general manipulator equation:
	\begin{align}
		M(q) \genAcc + C(\genPos, \change{\genVel}) &= g(\genPos) + Bu + \contactJacobian(\genPos)^T \lambda,
		\label{eq:dynamics}
	\end{align}
	where $M \in \Real^{n_v \times n_v}$ is the mass matrix, $C \in \Real^{n_v}$ and $g \in \Real^{n_v}$ are the Coriolis and gravitational forces respectively, $B \in \Real^{n_v \times n_u}$ is the actuator matrix, $u \in \Real^{n_u}$ is the vector of actuator inputs, and $\contactJacobian \in \Real^{n_v \times n_c}$ and $\lambda \in \Real^{n_c}$ are the Jacobian of the active contact and holonomic constraints and the corresponding constraint forces respectively.
	\change{
		Note that $M(q)$ and $\contactJacobian(q)$ only have a dependence on $q$. We omit these dependencies for the remainder of this paper for conciseness.
	}

	\subsection{Rigid Body Impacts}
	
	In this paper, we model the complex deformations and surface forces that occur when a legged robot makes contact with a surface using a rigid-body contact model.
	This contact model does not allow deformations; instead, impacts are resolved instantaneously.
	Therefore, the configuration remains constant over the impact event, all forces except for the contact forces are considered negligible, and the velocities change instantaneously according to the contact impulse $\contactImpulse$:
	\begin{align}
		M(\genVelPost - \genVelPre) &= \contactJacobian^T \contactImpulse,
		\label{eq:impact_map}
	\end{align}
	where $\genVelPre$ and $\genVelPost$  are the pre- and post-impact velocities respectively, $\contactJacobian$ is the Jacobian for the active constraints of the new contact mode, and $\contactImpulse$ is the impulse sustained over the impact event.
	If we include the standard constraint that the new stance foot does not move once in contact with the ground (purely inelastic collision and no-slip condition),
	\begin{align}
		\contactJacobian \genVelPost = 0,
	\end{align}
	\noindent $\contactImpulse$ can be solved for explicitly, determining the post-impact state $\genStatePost$ purely as a function of the pre-impact state $\genStatePre$:
	\begin{align}
		\genPosPost &= \genPosPre \label{eq:pos_reset_map},\\
		\genVelPost &= (I - M^{-1} \contactJacobian^T (\contactJacobian M^{-1} \contactJacobian^T)^{-1} \contactJacobian) \genVelPre \label{eq:vel_reset_map}.
	\end{align}
	\change{This reset map is commonly enforced as a constraint between hybrid modes separated by an impact event when modeling the hybrid dynamics of legged robots \cite{reher2021inverse} \cite{li2020hybrid}.}

	\subsection{Operational Space Control}
	We use an operational space controller (OSC) to track and stabilize the reference trajectories for the various motions explored in this paper.
	An OSC is a model-based inverse dynamics controller that tracks a general set of task or output space accelerations by solving for dynamically consistent inputs, ground reaction forces, and generalized accelerations \cite{wensing2013generation}
	\cite{sentis2005control}.
	For an output position $y = \phi(q)$ and corresponding output velocity $\dot y = J_y(q) \genVel$, where $J_y(q) = \PartialDiff{\phi}{q}$,
	the commanded output accelerations $\ddot {y}_{cmd}$ are calculated from the feedforward reference accelerations $\ddot y_{des}$ with PD feedback:
	\begin{align}
		\ddot {y}_{cmd} &=\taskSpaceAccDes + K_p (\taskSpacePosDes - \taskSpacePos) + K_d (\taskSpaceVelDes - \taskSpaceVel),
		\label{eq:output_cmd}
	\end{align}
	\change{where $K_p$ and $K_d$ are feedback gains on the task space position and velocity error respectively.}
	The objective of the OSC is then to produce dynamically feasible output accelerations $\ddot y$ given by: 
	\begin{align*}
		\ddot y &= \dot J_y \genVel + J_y \genAcc,
	\end{align*}
	such that the instantaneous output accelerations of the robot are as close to the commanded output accelerations as possible.
	This controller objective can be nicely formulated as a quadratic program (QP):
	
	\change{
		\begin{align}
			\min_{u, \lambda, \genAcc}   && \sum_{i}^{N} \Norm{\tilde{\taskSpaceAcc}_i}_{W_i}  + \Norm{u}^2_{W_u} +  \Norm{\genAcc}^2_{W_{acc}} + \Norm{\lambda}^2_{W_{\lambda}} && & \label{eq:osc_qp_full}\\
			\text{s.t.}
			&&  M \genAcc  + C = g + Bu + \contactJacobian^T \lambda  &&  & \label{eq:dyn_constraint_full} \\ 
			&&  \contactJacobian \genAcc + \dot \contactJacobian \genVel = 0   && & \label{eq:holonomic_constraint_full} \\ 
			&& u_{min} \leq u \leq u_{max} && & \label{eq:actuator_limits_upper} \\
			&& \mu \lambda_z \leq \lvert \lambda_x \rvert \label{eq:friction_cone_x}  &&  & \\
			&& \mu \lambda_z \leq \lvert \lambda_y \rvert \label{eq:friction_cone_y}  &&  & \\
			&& \lambda_n \geq 0,   &&  & \label{eq:friction_cone_constraint_full}	
		\end{align}
	}
	
	where $i$ denotes the particular output being tracked (e.g., center of mass or foot position) and $W_i$ are corresponding weights on tracking that output.
	The QP seeks to minimize the acceleration tracking error, $\tilde{\taskSpaceAcc}_i = \ddot y_{i_{cmd}} - \ddot y_i$, for the weighted sum across all tracking objectives. Additional regularization costs can be added to avoid non-unique solutions if the problem is underspecified. \cref{eq:dyn_constraint_full} is the dynamics constraint, where $\contactJacobian$ is the Jacobian for the active constraints for the current mode, and $\lambda$ are the corresponding constraint forces. \cref{eq:holonomic_constraint_full} enforces the holonomic constraints, and a linear approximation of the friction cone constraint is given by \cref{eq:friction_cone_x}, \cref{eq:friction_cone_y}, and \cref{eq:friction_cone_constraint_full}.
	Here, $\lambda_z$ is the normal component of the contact force and $\lambda_x, \lambda_y$ are the tangential components expressed in the robot frame.
	\change{
		$W_u$, $W_{acc}$, and $W_{\lambda}$ are general regularization weights on the decision variables, specifically the inputs, generalized accelerations, and contact forces respectively.
	}
	The QP can be solved quickly ($>$1000 Hz), even for complex robots such as Cassie.
	\section{Impact Invariance}
	\label{sec:impact_invariance}
	
	In this section, we present the key ideas of this work. 
	In \cref{subsec:challenge_of_control}, we explain the common challenges that arise when applying feedback during impacts.
	We then propose a solution in \cref{subsec:impact_invariant_subspace}, which also introduces the concept of the impact-invariant subspace.
	In \cref{subsec:implementation} and \cref{subsec:blend_projection}, we explain the practical details of how to implement the subspace in the context of an operational space controller.
	\change{
		Finally, in \cref{subsec:robust_control_formulation} we derive our formulation for impact-invariant control from the perspective of robust optimal control.
	}

	\subsection{Challenges of Control During Impacts}
	\label{subsec:challenge_of_control}
	To motivate the concept of the impact-invariant subspace, we begin by highlighting and describing the difficulties of applying feedback control during an impact event.
	For the sake of simplicity, we consider a feedback controller with constant feedback gains that controls an output $\taskSpacePos: \Real^{n_q} \rightarrow \Real^{n_y}$ to track a time-varying trajectory $\taskSpacePosDes(t):~ \left[0, \infty \right) \rightarrow \Real^{n_y}$ by driving the tracking error $\tilde{\taskSpacePos}(t) = \taskSpacePosDes(t) - \taskSpacePos(t)$ to zero.
	This is commonly accomplished with control law $\nominalControl = \feedforwardControlTerm + \feedbackControlTerm$ where $\feedforwardControlTerm$ is the feedforward controller effort required to follow the reference acceleration $\taskSpaceAccDes$ and $\feedbackControlTerm$ is the PD feedback component given by:
	\begin{align}
		u_{fb}(t) = K_p \tilde{y}(t) + K_d \dot{\tilde{y}}(t).
	\end{align}
	
	\begin{figure}[ht]
		\centering
		\includegraphics[width=0.48\textwidth]{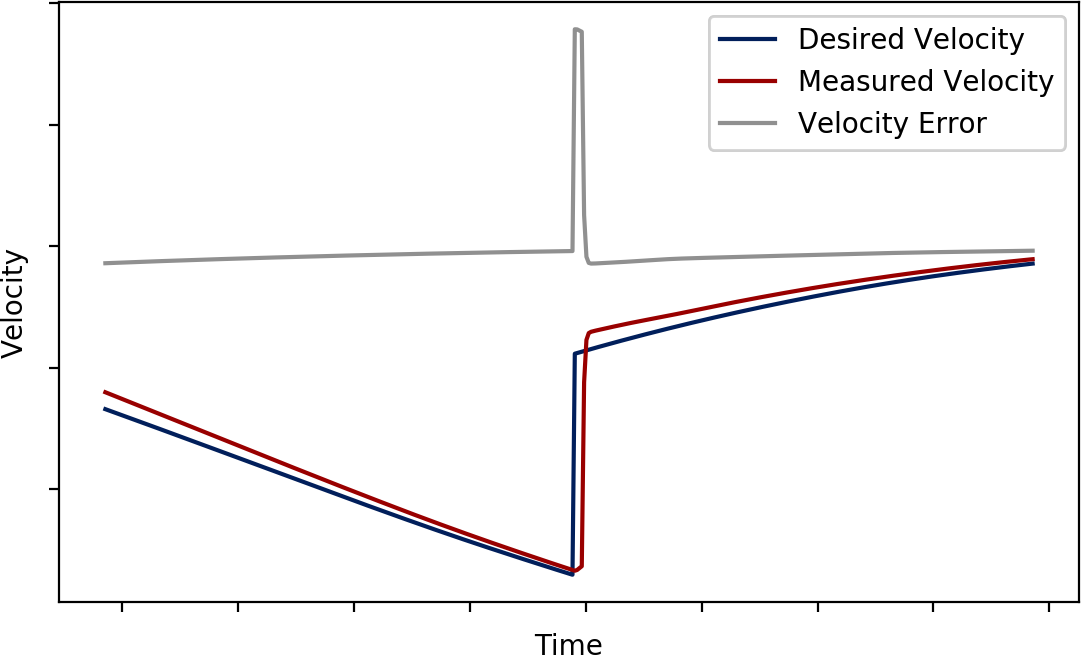}
		\caption{Illustration of a system that undergoes an impact event. The desired velocity plan correctly includes the discontinuity as predicted by rigid body impact laws and the measured velocity is being properly regulated to match the desired plan. However, due to the mismatch in impact time, the velocity error inevitably spikes during the impact event.}
		\label{fig:instantaneous_impacts}
	\end{figure}
	
	The reference trajectory $\dot{y}_{des}(t)$ for systems that make contact with their environment has discontinuities at the impact events in order to be dynamically consistent with \cref{eq:vel_reset_map}.
	Therefore, in a short time window around an impact event, there will be a discontinuity in the reference trajectory $\dot{y}_{des}(t)$ at the nominal impact time and a near-discontinuity in the actual robot state when the system $\dot{y}(t)$ makes contact with the ground as shown in \cref{fig:instantaneous_impacts}.
	Because the robot configuration is approximately constant over the impact event, the change in controller effort is governed by the change in velocity error:
	\change{
		\begin{align}
			\Delta u \approx K_d \Delta (\taskSpaceVelDes - \taskSpaceVel),
		\end{align}
	}
	thus any mismatches in impact timing will unavoidably result in spikes in the feedback error signal, which has been similarly noted in \cite{rijnen2017control}.
	
	\begin{remark}
		The previous example assumes the jump in the reference trajectory is time-based. Although it is possible to formulate trajectories with event-triggered jumps, these methods require detection, which for state-of-the-art methods still have delays of 4-5 ms \cite{bledt2018cheetah}. Moreover, in reality, impacts are not resolved instantaneously but rather over several milliseconds to tens of milliseconds. In this time span, it is not clear which reference trajectory to use, as using either trajectory will output a large tracking error.
	\end{remark}
	
	Note that a large tracking error, shown in \cref{fig:instantaneous_impacts}, results from only a small difference in impact timing, yet the controller will respond to the large velocity error and introduce controller-induced disturbances.
	Furthermore, this sensitivity to the impact event is amplified by the large contact forces that impair state estimation and result in likely inaccurate velocity measurements due to necessary filtering.
	
	The key challenges of applying feedback during impacts can thus be summarized as: impacts are brief moments of high uncertainty where our references are poorly defined and our measurements are inaccurate. 
	
	\begin{figure*}[ht]
		\begin{subfigure}[t]{0.32\textwidth}
			\includegraphics[width=1.0\textwidth]{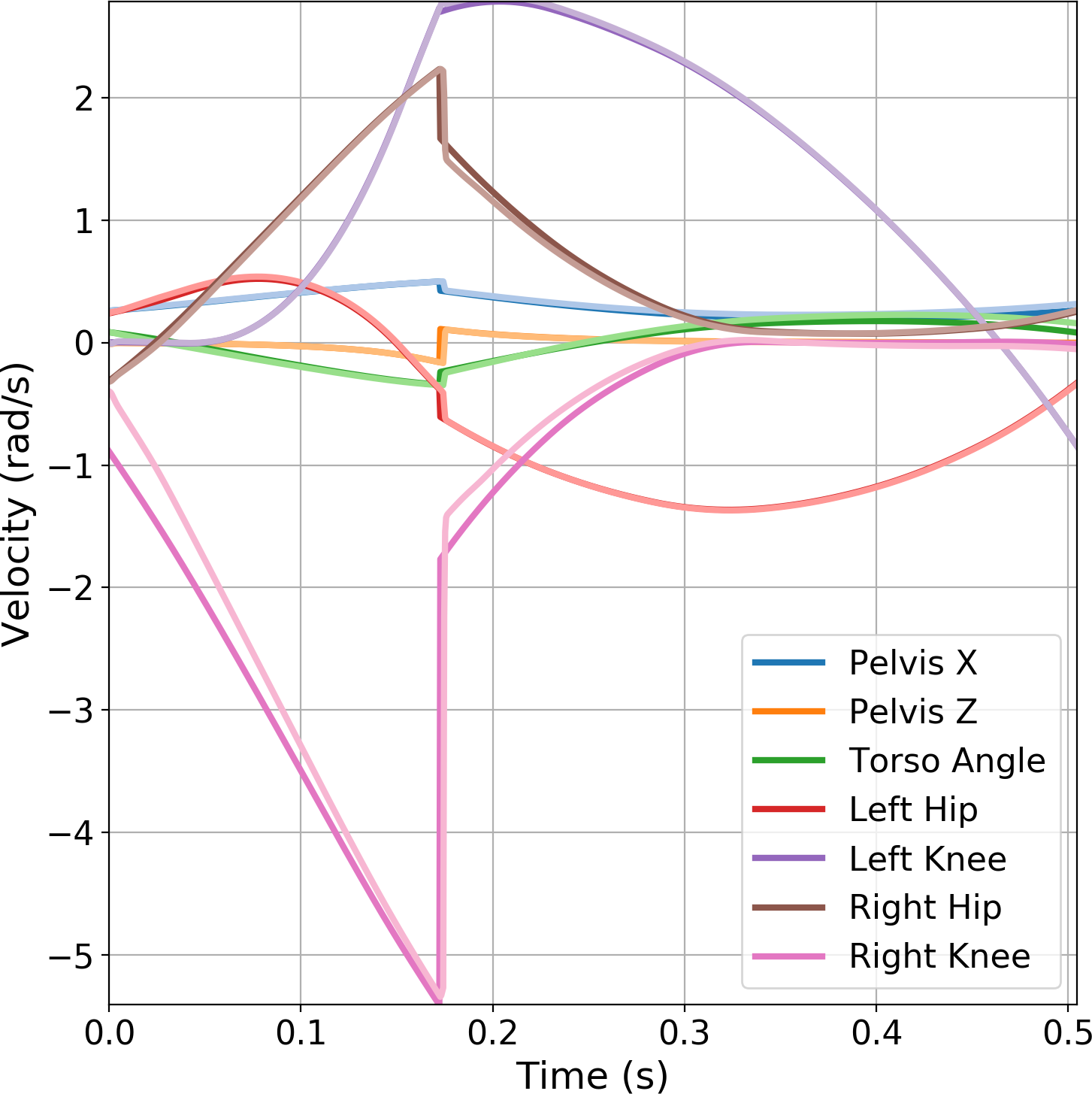}
			\caption{Generalized velocities for a periodic walking trajectory for a planar five-link biped.}
			\label{subfig:rabbit_gen_vel}
		\end{subfigure}
		\begin{subfigure}[t]{0.32\textwidth}
			\includegraphics[width=1.0\textwidth]{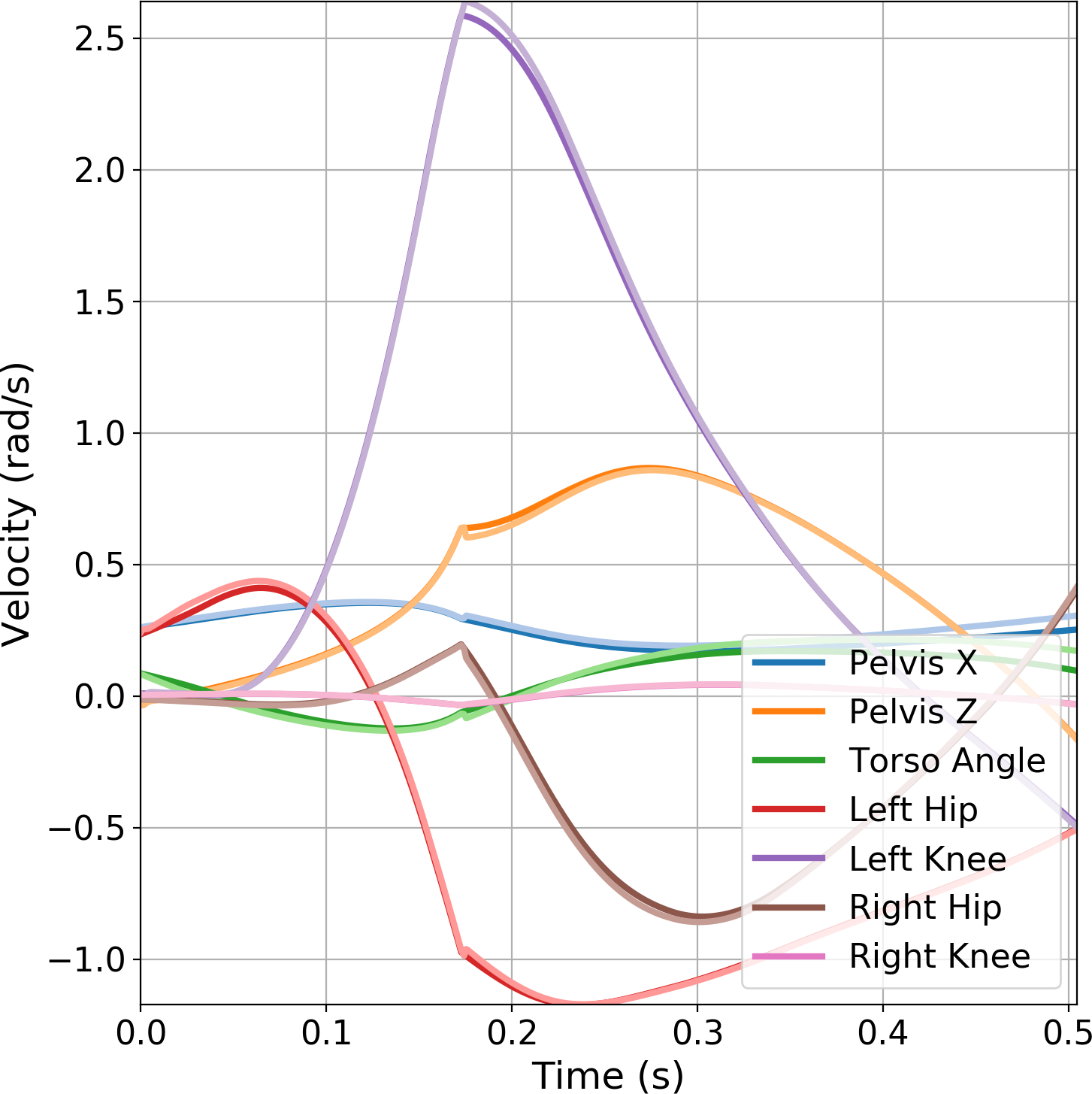}
			\caption{Generalized velocities projected to the impact-invariant space}
			\label{subfig:rabbit_proj_vel}
		\end{subfigure}
		\begin{subfigure}[t]{0.32\textwidth}
			\includegraphics[width=1.0\textwidth]{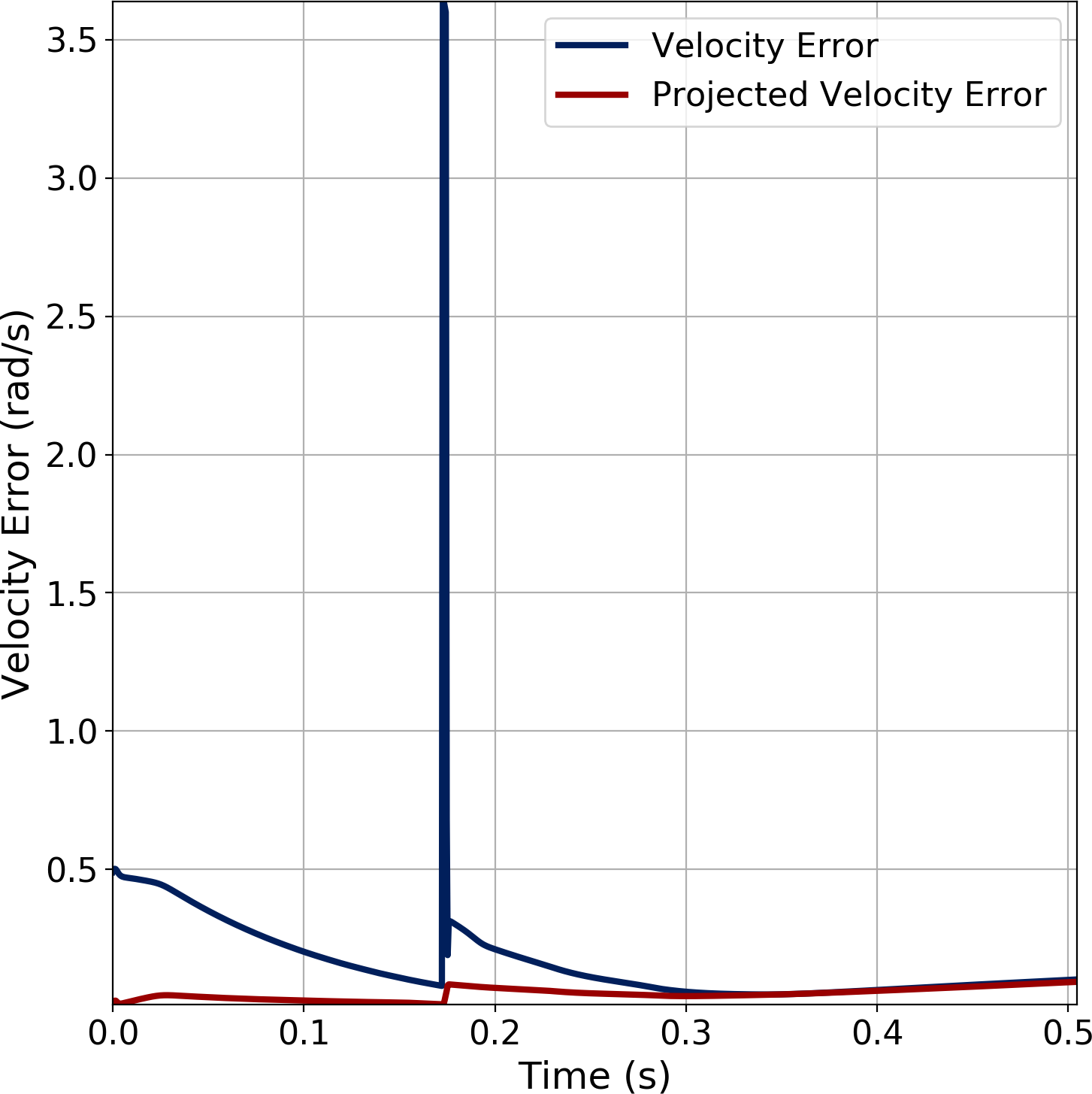}
			\caption{Comparison of tracking errors in the two spaces}
			\label{subfig:rabbit_vel_error}
		\end{subfigure}
		\caption{\change{The velocities are shown for a periodic symmetric walking trajectory (a) for a planar five-link biped. Because the trajectory is constructed to be symmetric, only half of the gait is shown without loss of information. The instantaneous jump in velocities is due to an impact event when the right foot makes contact with the ground. The darker shade indicates the nominal trajectory while the lighter shade is an example of actual velocities when tracking to the nominal trajectory. Despite the qualitatively ``good" tracking of the velocities, the error near the impact event still exhibits a sharp jump (c) due to the discontinuity from impact. The same velocities projected to the impact-invariant subspace (b), do not have experience this discontinuity and therefore results in a much smaller tracking error, which is a better reflection of the actual system tracking.}}
		\label{fig:impact_invariant_example}
	\end{figure*}
	
	\subsection{Impact-Invariant Subspace}
	\label{subsec:impact_invariant_subspace}
	The key insight in resolving the problem of control during impacts is inspired by \cite{gong2022zero}, in which Gong and Grizzle delineate desirable properties of angular momentum about the contact point.
	They highlight that it is invariant over impacts on flat ground, meaning that it is continuous over the impact event despite it being a function of velocity.
	
	The concept of an impact-invariant subspace is a generalization of this property.
	We observe that there is a space of velocities that, like angular momentum about the contact point, are continuous through impacts for \textit{any} contact impulse.
	By switching to track only these outputs in a small time window around anticipated impacts, we avoid controller-induced disturbances from uncertainty in the impact event.
	While angular momentum is \change{in} $\Real^3$ or $\Real^2$ for planar systems such as Rabbit, the impact-invariant subspace is in $\Real^{n_v - n_c}$, where $n_v$ is the dimension of generalized velocities and $n_c$ is the dimension of the contact impulse of the impact event.
	For Rabbit walking, this space is in $\Real^{7-2}$.
	For Cassie, each foot on the ground imposes a contact constraint of dimension 5 and the four bar linkage on each leg provides a distance constraint of dimension 1 that is always active\endnote{We model Cassie's feet as two point contacts on the same rigid body, with each point imposing a constraint of dimension 3. One dimension is redundant, thus resulting in an overall active contact constraint dimension of 5.}.
	Therefore the impact-invariant subspace is in $\Real^{18 - 7}$ for impacts with a single foot (walking, running) and is in $\Real^{18 - 12}$ for impacts with both feet (jumping).
	The implication of the impact-invariant subspace is a space that provides full robustness to uncertainty caused by impacts, while minimally reducing control authority. This results in better tracking which is important for precise dynamic motions.

	The impact-invariant subspace is defined as the nullspace of $(M^{-1} \contactJacobian^T)^T$ or left nullspace \cite{strang1993introduction} of $M^{-1} \contactJacobian^T$, where $\contactJacobian$ again is the Jacobian for the active constraints.
	Thus a basis $P(q) \in \Real^{(n_v - n_c) \times n_q}$ for this nullspace is such that:
	\change{
		\begin{align}
			P M^{-1} \contactJacobian^T \Lambda = 0.
			\label{eq:ii_projection}
		\end{align}
	}
	\change{
		The velocities during the impact event can be described as $\genVel = \genVelPre + M^{-1} \contactJacobian^T \Lambda$, where here $\Lambda$ represents the contact impulse experienced thus far.
		The velocity during impact in this basis remains the same as the pre-impact velocity, $P \genVel = P (\genVelPre + M^{-1} \contactJacobian^T \Lambda) = P (\genVelPre$), which is the intended effect. 
		That is, for any contact impulse $\Lambda$, the velocities in the impact-invariant subspace are unchanged.
		To project the generalized velocities down to the impact-invariant subspace, we can create an orthonormal projection matrix $Q(q) \in \Real^{n_v \times n_v} = P^T P$. 
		In practice, we can compute this projection matrix as $Q = I - M^{-1} \contactJacobian^T (\contactJacobian M^{-T} M^{-1} \contactJacobian^T) \contactJacobian M^{-T}$, which we use to define the projected velocity error as $Q(\genVel_{des} - \genVel)$.
	}
	
	\change{
		Let's consider a concrete example of how the impact-invariant subspace can be applied to control of a legged robot using a walking gait of a simulated five-link biped.
		We solve a hybrid trajectory optimization problem  \cite{posa2016optimization} to generate a periodic walking gait for the planar biped that satisfies the rigid body impact law \cref{eq:impact_map}.
		The generalized velocities for half of gait are shown in \cref{subfig:rabbit_gen_vel}, where the jump in velocities is a result from the right foot of the robot making impact with the ground.
		As noted in the figure caption, the darker shade indicates the nominal or target velocities, while the lighter shade indicates an example of velocities the robot may experience when tracking to the nominal trajectory.
		For these same velocities, we can compute and apply the impact-invariant projection to arrive at the projected velocities shown in \cref{subfig:rabbit_proj_vel}.
		An important observation of the projected velocities is that they are now continuous over the impact event, which eliminates the large spike in tracking error that was a result of tracking a discontinuous signal as shown in \cref{subfig:rabbit_vel_error}.
		Notice, although every degree of freedom experiences a jump in its velocities at the impact event, the projected velocities not zero. 
		In fact, the original and projected velocities of the left knee joint are remarkably similar, which can be intuitively understood by the fact that forces that left knee is only distantly connected to forces that enter through the right foot according to the kinematic tree.
		In contrast, the projected velocities of the right knee joint are practically zero.
	}
	
\begin{figure}
		\centering
		\includegraphics[width=0.48\textwidth]{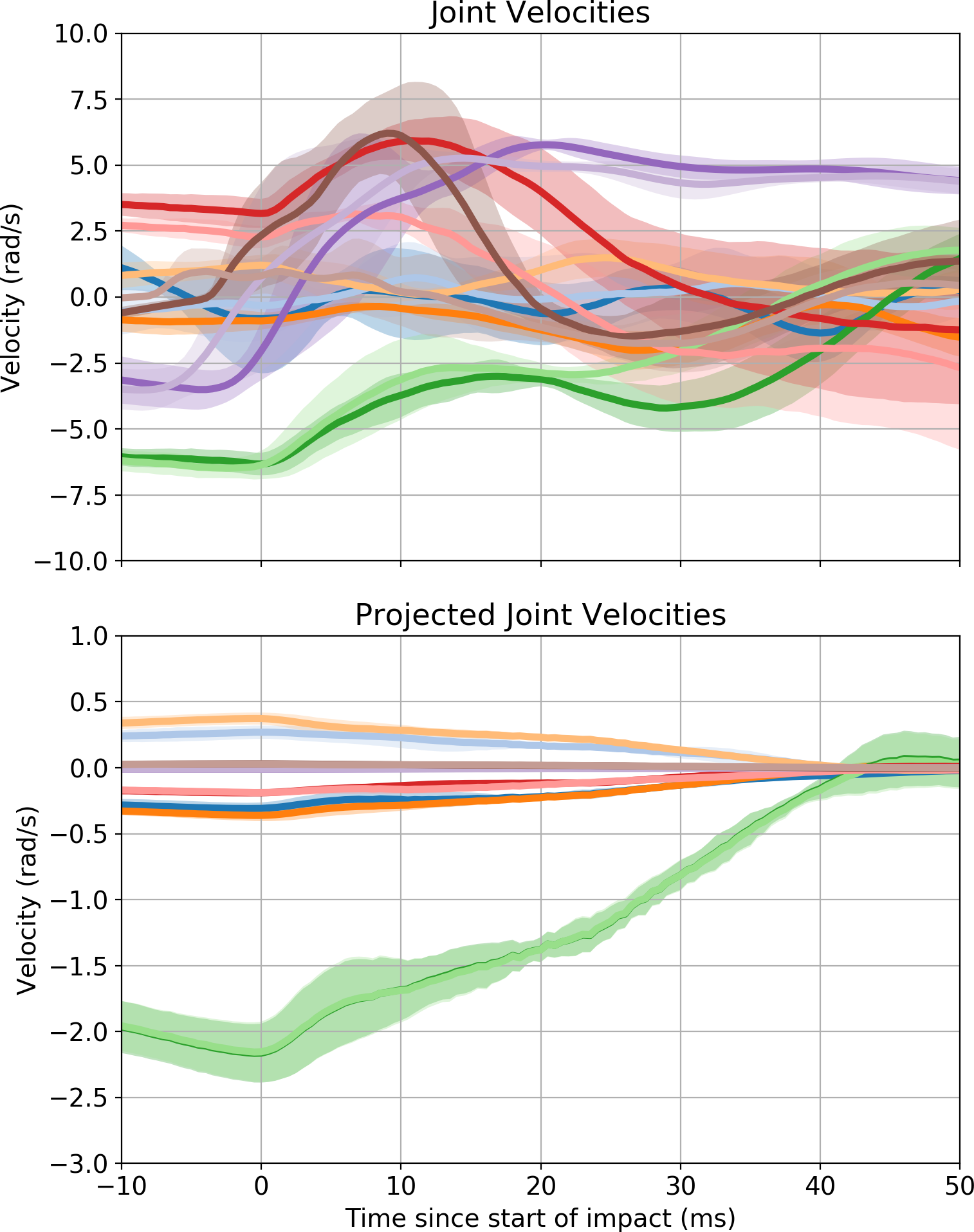}
		\includegraphics[width=0.40\textwidth]{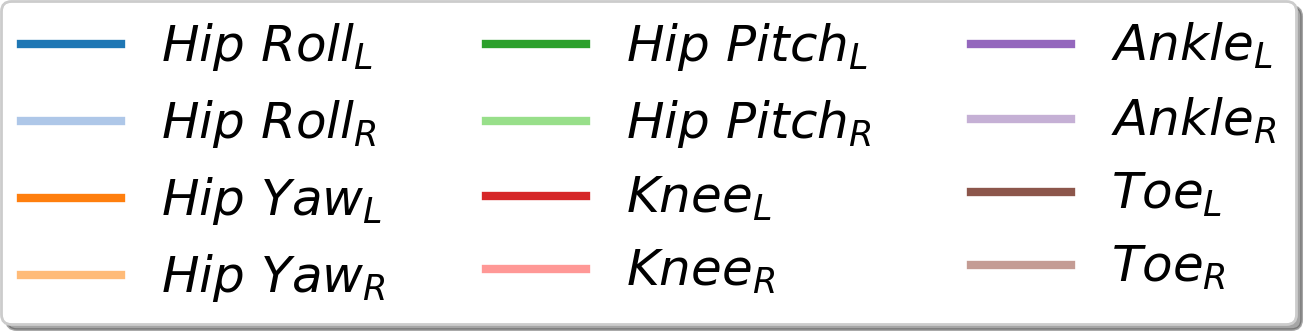}
		\caption{Demonstration of the impact-invariant projection on joint velocity data from 8 consecutive jumping experiments on the physical Cassie robot. Joint velocities (top) during the landing event change rapidly. By projecting the same joint velocities to the impact-invariant subspace (bottom), the values are more consistent and more amenable for feedback control. Note, the change in joint velocities primarily occurs within a time span of only 10 - 20 ms. The L and R subscripts indicate the left and right leg respectively.}
		\label{fig:hardware_joint_vels}
	\end{figure}

	Similarly, to illustrate the benefit on a \textit{physical robot}, the joint velocities for Cassie executing a jumping motion  right when it lands \change{across 8 experiments} are shown in \cref{fig:hardware_joint_vels}. Details of the jumping controller and experiments are given in \cref{sec:jumping_controller}.
	Observe that the projected joint velocities are significantly smoother than the original joint velocities.
	
	\subsection{Implementation for Task Space Tracking}
	\label{subsec:implementation}
	
	The objective of impact-invariant control is to apply control on the subspace where the changes in the velocity introduced via the impact map, $M^{-1} \contactJacobian^T$, do not appear.
	In practice, we accomplish this by modifying our controllers to eliminate the component of the velocity error that is within the subspace spanned by the impact map $M^{-1} \contactJacobian^T$ by solving the following optimization problem:
	\begin{align}
		\min_{\lambda} &\quad& \TwoNorm{\taskSpaceVelDes - \taskSpaceJacobian(\genVel + M^{-1} \contactJacobian^T \lambda)}
		\label{eq:ii_optimization_problem}\\
		\text{s.t.}
		&\quad& \holonomicJacobian (\genVel + M^{-1} \contactJacobian^T \lambda) = 0. \label{eq:constraint_vel} 	
	\end{align}
	\change{Here, we overload $J_y$ to represent the matrix constructed by stacking the active output space Jacobians $J_{y,i} \forall i$.}
	$J_h$ is the Jacobian for the holonomic constraints that are unambiguously active during impact\endnote{Unambiguously active constraints refer to constraints that are active both before and after the impact event. For example, the four-bar linkage constraint is always active and for walking with a double stance phase, the contact constraint for the current stance foot is active both before and after the other foot makes contact. To be explicit, $J_h$ is a subset of $\contactJacobian$ because $\contactJacobian$ includes all active constraints of the current impact event.}.
	
	This reduces the total error by subtracting a correction term $M^{-1} \contactJacobian^T \lambda$ that by construction is in the range of $M^{-1} \contactJacobian^T$.
	We must include the constraint forces for all active constraints, as contact forces from the impact event will cause corresponding reaction forces, which should still satisfy kinematic feasibility of the velocities enforced by \cref{eq:constraint_vel}. 
	Notice that this optimization problem is an equality constrained QP, which we are able to solve this in closed form:
	\begin{align}
		\begin{bmatrix}
			\lambda^* \\
			\mu^*
		\end{bmatrix} = \begin{bmatrix}
			A^T A & B^T \\
			B & 0 
		\end{bmatrix}^{-1} \begin{bmatrix}
			A^T (\taskSpaceVelDes - \taskSpaceVel) \\ J_h \genVel
		\end{bmatrix} \label{eq:kkt_closed_form},
	\end{align}
	where $A = J_y M^{-1} \contactJacobian^T$, $B = J_h M^{-1} \contactJacobian^T$, and $\mu^*$ is the Lagrange multiplier for the holonomic constraint.
	
	Applying $\lambda^*$ back into the OSC formulation, we combine the correction and the measured velocity to define the projected generalized velocities as:
	\begin{align}
		\projVel &= \genVel + M^{-1} J_{\lambda}^T \lambda^*.
		\label{eq:vel_proj}
	\end{align}
	We then define the projected output velocity, $\dot{y}_{proj}$, and projected output space velocity error, $\projTaskSpaceVelError$, as:
	\begin{align}
		\dot{y}_{proj} &= J_y \projVel 
		\label{eq:osc_ii_velocity} \\
		\projTaskSpaceVelError &= \taskSpaceVelDes - \projTaskSpaceVel.
		\label{eq:ii_vel_error}
	\end{align}

	\change{
		The constrained least-squares problem \cref{eq:ii_optimization_problem} used to modify any general task-space velocity error to be impact-invariant is a generalization of the projection matrix $Q$ from \cref{eq:ii_projection}.
		We show that \cref{eq:ii_optimization_problem} performs the same modification to the tracking error as $Q$ in \cref{sec:appendices}.
		The projected output space velocity error $\projTaskSpaceVelError$ is then used in the OSC feedback law (\cref{eq:output_cmd}) to create the impact-invariant feedback law:
		\begin{align}
			\ddot {y}_{cmd} &= \taskSpaceAccDes + K_p (\taskSpacePosDes - \taskSpacePos) + K_d (\taskSpaceVelDes - \projTaskSpaceVel).
			\label{eq:ii_feedback_law}
		\end{align}
	}
	
	\change{
		Stacking the output space Jacobians to construct a single $J_y$ ensures only a single $\lambda^*$ is used to project the tracking error.
		Note, the derivative gains and weighting matrices defined in \eqref{eq:output_cmd} \eqref{eq:osc_qp_full} can intuitively be included when constructing $A$ in \eqref{eq:kkt_closed_form}, but in practice we did not find a noticeable effect from including them.
	}
	
	\subsubsection{Constraints on the Projection}
	
	\change{
		Due to the lack of constraints on $\lambda$, the projection impulse is not guaranteed to be physically possible.
		$\lambda$ could be constrained to lie within the friction cone $FC$.
		However, upon further examination, inclusion of these constraints may be undesirable.
		This is because sensitivity to the impact event can result from the \textit{absence} of expected impacts as well - consider the case when the robot makes contact after the nominal impact time.
		Constraining $\lambda \in FC \cup -FC$ is not practical as this set is non-convex.
		Although it is possible to formulate this as a binary mixed integer program, with a integer variable per point contact, solving this problem was considered to require too many assumptions to justify the additional complexity.
	}
	
	\subsection{Activating the Projection}
	\label{subsec:blend_projection}
	
\begin{figure}
		\centering
		\includegraphics[width=0.48\textwidth]{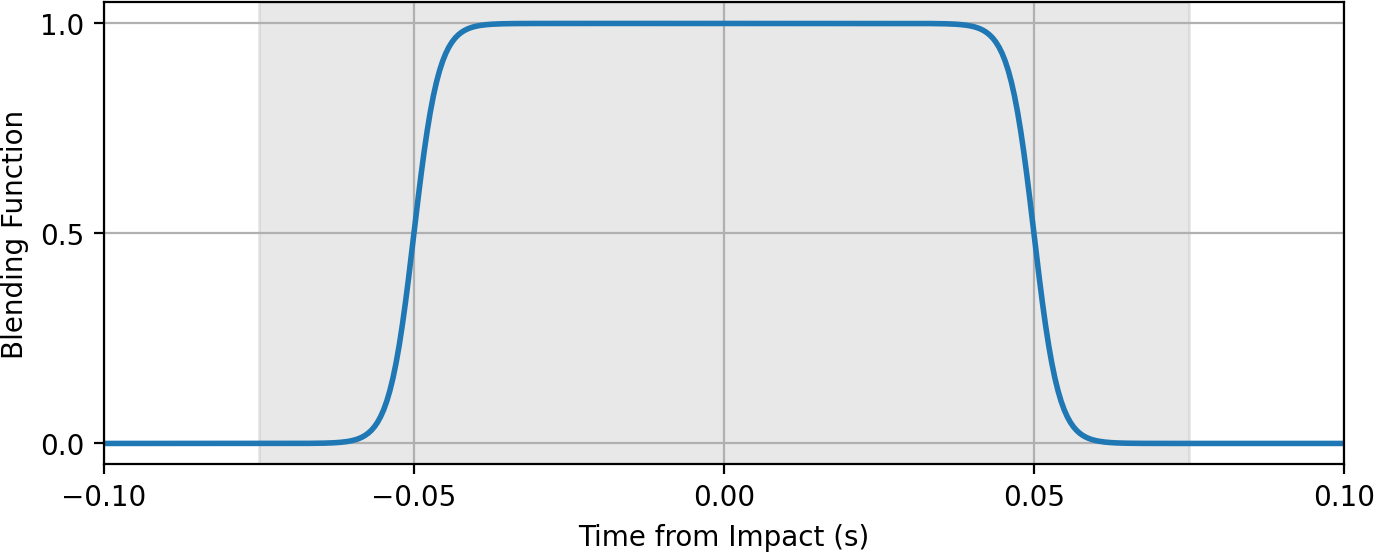}
		\caption{Blending function for the impact-invariant projection for a window $T$ of 50 ms and time constant $\tau$ of 2 ms.}
		\label{fig:blend_function}
	\end{figure}

	\change{
		The benefit of explicitly computing the error correction $M^{-1} \contactJacobian^T \lambda^*$ is that it becomes trivial to smoothly blend in the projection.
	}
	Because impact-invariant control essentially ignores errors in the space of impacts, we should only use it when we anticipate impacts.
	In practice, we do this by only activating the projection in a time window near anticipated impact events. 
	To avoid introducing discontinuities when activating the projection, we blend in the correction, $M^{-1} J_{\lambda}^T \lambda^*$ using a blending function \change{ $\alpha(t)$ visualized in \cref{fig:blend_function} constructed using a sigmoid function $\sigma(x) = \frac{e^{x}}{1 + e^{x}}$.
		\begin{align}
			\alpha(t) = \begin{cases} 
				\sigma(\frac{T - \lvert t - t_{s} \rvert}{\tau}) & t_{s} - 1.5T \leq t \leq t_{s} + 1.5T\\
				0 & otherwise
			\end{cases}
		\end{align}
		where $t$ is the current time, $t_{s}$ is the switching time defined as the nominal impact time of the reference trajectory, $T$ is the duration of the projection window, and $\tau$ is the time constant that determines the rate of the blending.
	}
	We then use $\alpha(t)$ to modify \eqref{eq:vel_proj} to be:
	\begin{align}
		\projVel &= \genVel + \alpha(t) M^{-1} J_{\lambda}^T \lambda^*.
	\end{align}
	Note our choice for the blending function $\alpha(t)$ is arbitrary; any monotonic continuous function with a range of $[0, 1]$ can accomplish a similar purpose.
	Additionally, while $\alpha(t)$ is a function of time, we can also activate/blend in the impact-invariant projection purely as a function of the robot's state.
	For example, we can use the distance between the foot or end effector and the environment in place of $t - t_{s}$.
	
	\change{
		\subsection{Robust Optimal Control Perspective}
		\label{subsec:robust_control_formulation}
	}
	\change{
		We can view impact-invariant control from the perspective of robust control, which formulates a min-max optimization problem \cite{salmon1968} that minimizes a control objective with subject to worst-case disturbances.
		Our control objective is to minimally perturb our control inputs, $\nu$, as is commonly done in the control barrier functions \cite{ames2019control}, where we take these inputs to be an intermediary value: the tracking error of our tracking objectives $\nu^* = \dot{y}_{des} - J_y v$, where $\nu \in \Real^{n_y}$.
		The disturbance is the unknown contact impulse $\Lambda$, which causes the measured velocity $\hat{v}$ to differ from the true velocity according to $\hat{v} = v + M^{-1} \contactJacobian^T \Lambda$.
		Because we can only measure $\hat{v}$, our control input is:
		\begin{align}
			\nu = \dot{y}_{des} - J_y (v + M^{-1} \contactJacobian^T \Lambda). \label{eq:measured_control_input}
		\end{align}
		However, since $\Lambda$ is unknown and potentially quite large, the controller \cref{eq:measured_control_input} is brittle.
		We formulate the following minimally perturbing robust controller, where robustness to arbitrary $\Lambda$ enters as the constraint that the controller must not be sensitive to $\Lambda$:
		\begin{align}
			\min_{\nu} \max_{\Lambda} &\quad & \Norm{\nu - \nu^*}_2^2 \label{eq:min_perturbation} \\
			\text{s.t.} &\quad & \frac{\partial \nu}{\partial \Lambda} = 0 \label{eq:sensitivity_constraint}.
		\end{align}
		If we restrict $\nu$ to be a linear function of the estimate $\nu(\hat{v}) = \dot y_{des} - J_y \hat{v}$, we reparameterize $\nu(\hat{v})$ as $\nu(\hat{v}) = Q (\dot y_{des} - J_y (\hat v))$, where $Q \in \Real^{n_y \times n_y}$.
		Under this parametrization, the robustness constraint can be simplified to $Q J_y M^{-1} \contactJacobian^T = 0$ and we arrive at the following matrix optimization problem:
		\begin{align*}
			\min_Q &\quad & \Norm{(Q - I) (\dot y_{des} - J_y v)}_2^2 \\
			\text{s.t.} &\quad & Q J_y M^{-1} \contactJacobian^T = 0.
		\end{align*}
	}
	
	\change{
		We cannot measure the true velocity $v$ in the expression $\dot y_{des} - J_y v$, so we instead minimize the matrix norm induced by the vector 2-norm, which is $\Norm{Q - I}_2 = \sup_{\dot y_{des} - J_y v \neq 0} \frac{\Norm{(Q - I)(\dot y_{des} - J_y v)}_2}{\Norm{\dot y_{des} - J_y v}_2}$.
		Minimizing this matrix norm is a semi-definite program (SDP) \cite{boyd2004convex}.
		It is not currently feasible to solve this SDP at real-time control rates, so we instead minimize an upper bound on the robust minimum-perturbation cost. 
		We substitute the 2-norm with a Frobenius norm and arrive at:
		\begin{align}
			\min_Q &\quad & \Norm{Q - I}_F^2 \nonumber \\
			\text{s.t.} &\quad & Q J_y M^{-1} \contactJacobian^T = 0,
			\label{eq:minimum_perturbation_qp}
		\end{align}
		which is equivalent to solving for a projection matrix for the left-nullspace of $J_y M^{-1} \contactJacobian^T$, which we show is equivalent to \cref{eq:ii_optimization_problem} in \cref{subsec:least_squares_and_projections}.
		Thus we have shown that projecting the velocities to the null-space of the impact map is the optimal linear policy, as formulated by \cref{eq:min_perturbation,eq:sensitivity_constraint}.
		Note, in place of using velocity tracking error as our control inputs, we could instead consider the final control effort, $\nu = u$, as in \cite{ames2019control}, for example.
		The projection would then depend on the nominal controller itself, via gains $K_d$ and weights $W_i$ as well as the system dynamics \cref{eq:dyn_constraint_full}.
		These scaling terms may slightly modify the solution, but as mentioned in \cref{subsec:implementation}, we did not find a noticeable effect in simulation experiments.
	}
	
	\section{Experiments and Evaluation}
	\label{sec:evaluation}
	
	\begin{figure*}[ht]
		\centering
		\begin{subfigure}[b]{0.19\textwidth}
			\includegraphics[height=0.11\textheight]{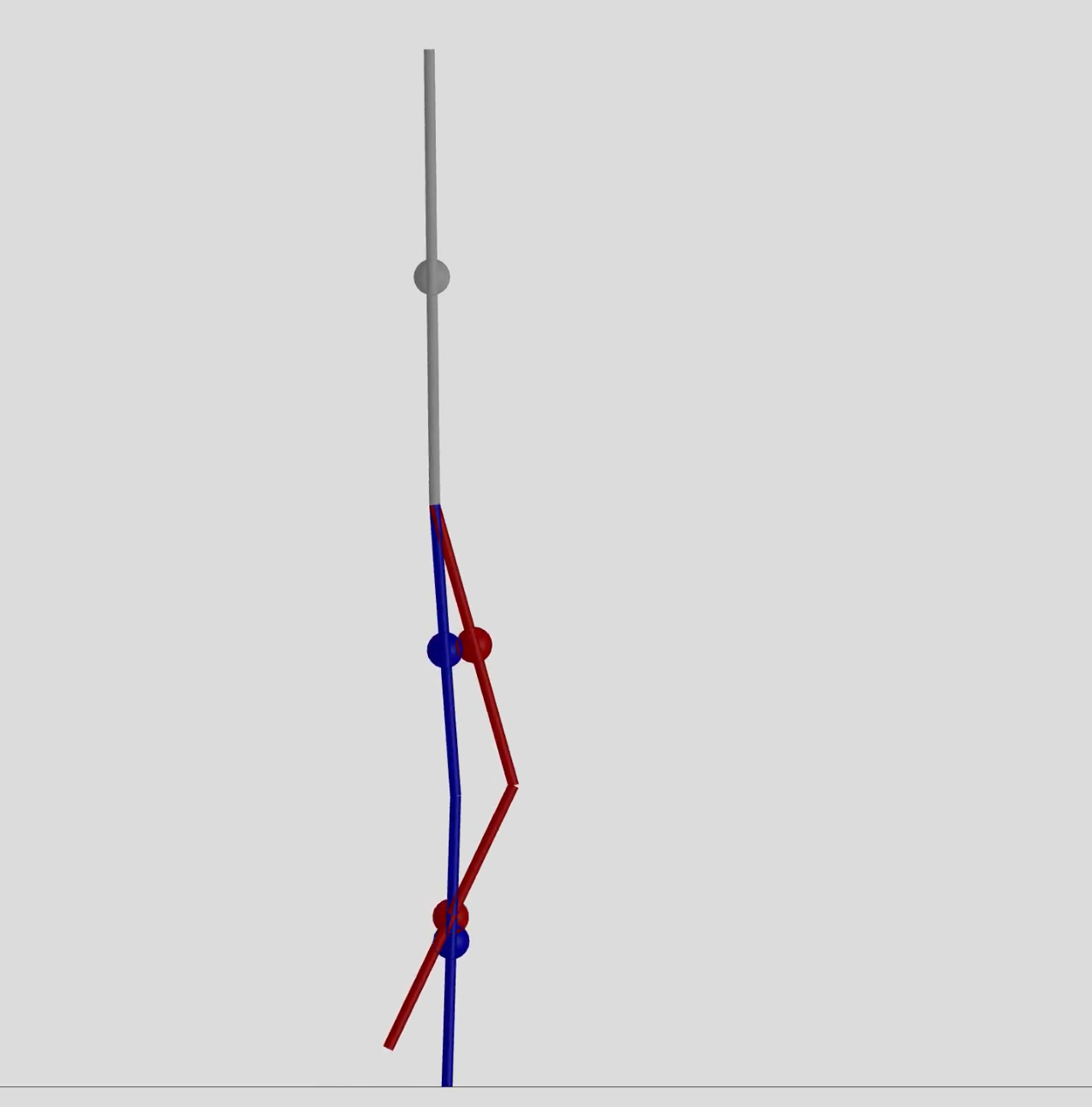}
			\caption{Walk}
		\end{subfigure}
		\begin{subfigure}[b]{0.178\textwidth}
			\includegraphics[height=0.11\textheight]{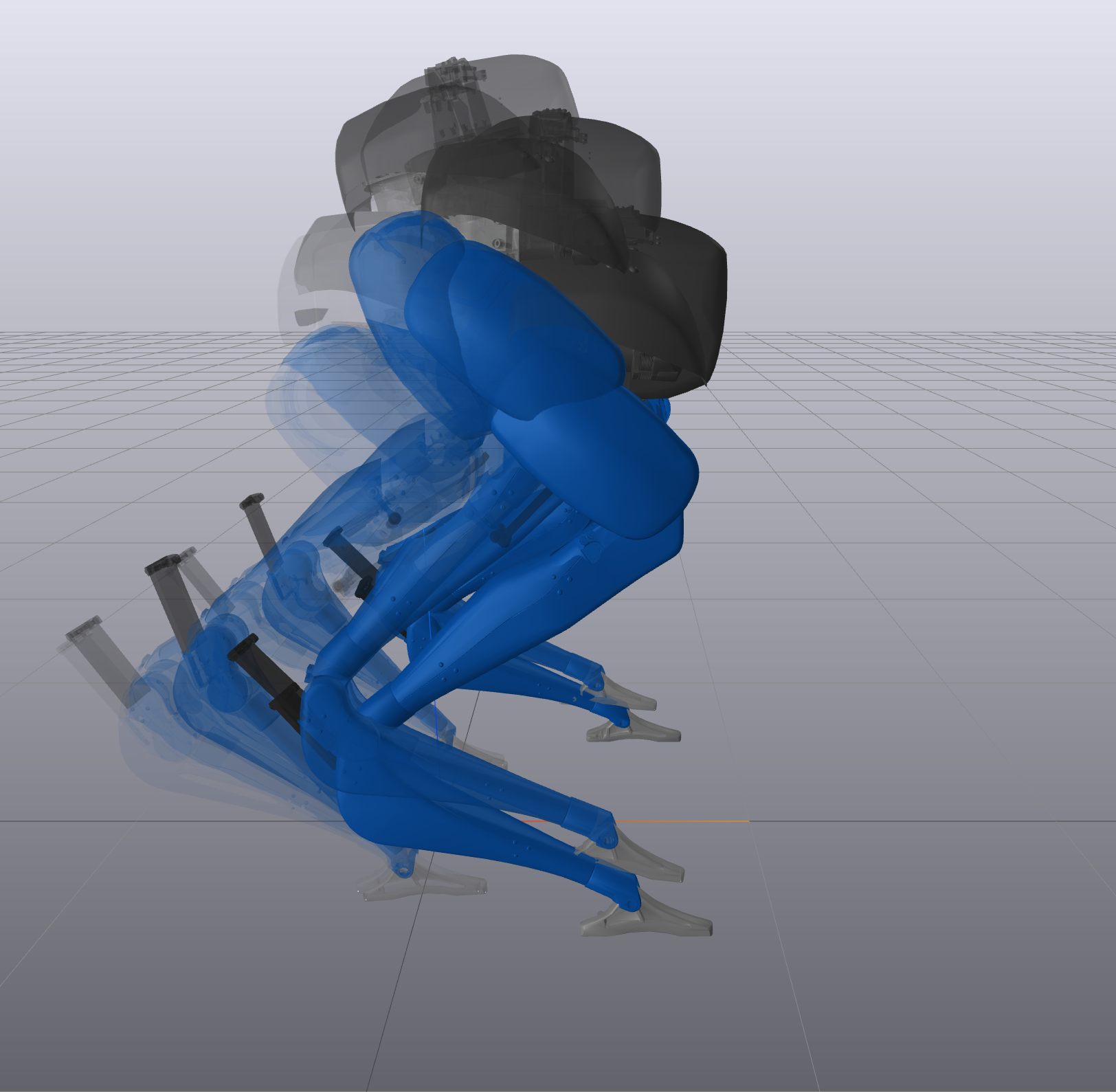}
			\caption{Jump}
		\end{subfigure}
		\begin{subfigure}[b]{0.24\textwidth}
			\includegraphics[height=0.11\textheight]{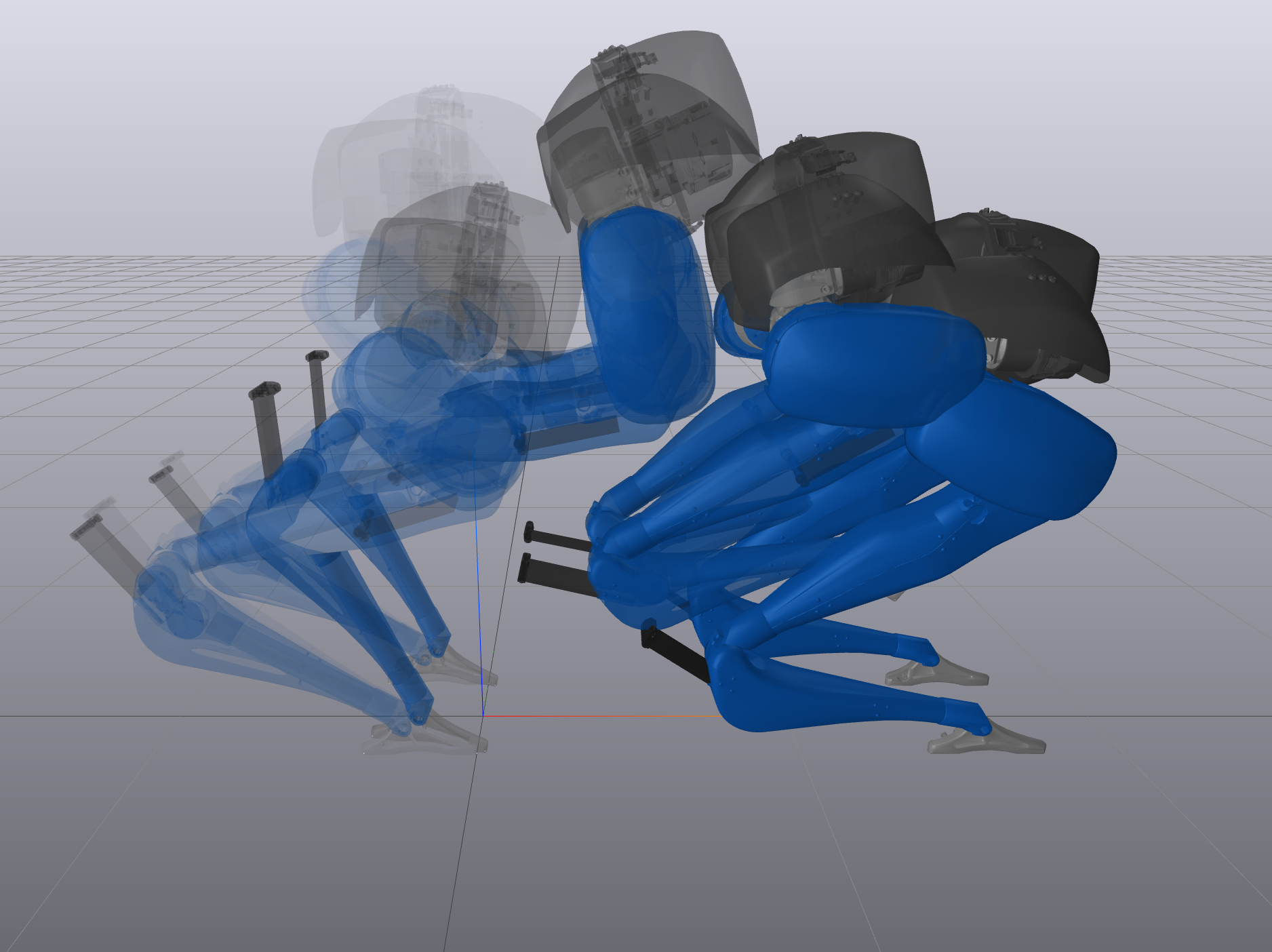}
			\caption{Long jump}
		\end{subfigure}
		\begin{subfigure}[b]{0.18\textwidth}
			\includegraphics[height=0.11\textheight]{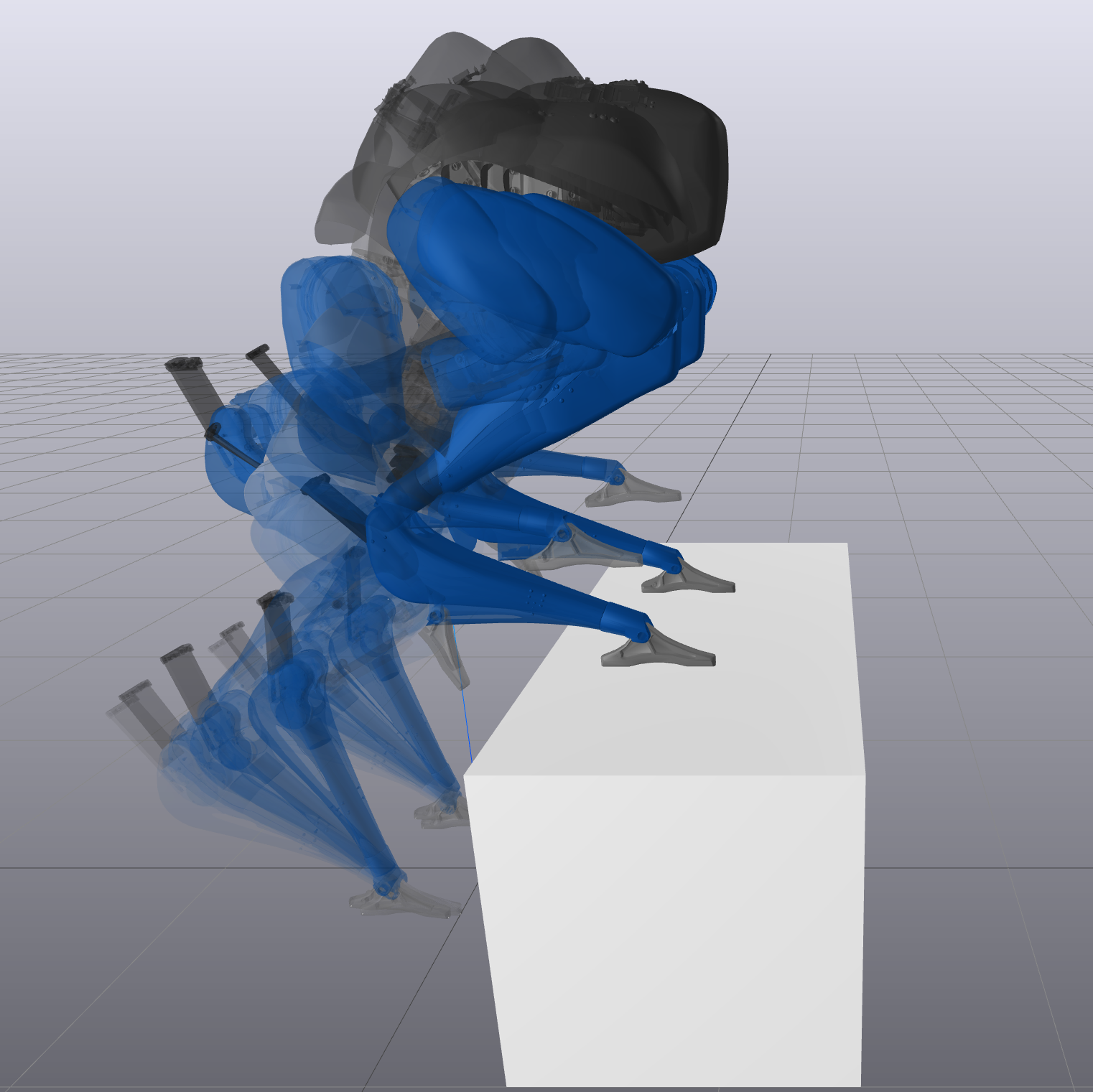}
			\caption{Box jump}
		\end{subfigure}
		\begin{subfigure}[b]{0.16\textwidth}
			\includegraphics[height=0.11\textheight]{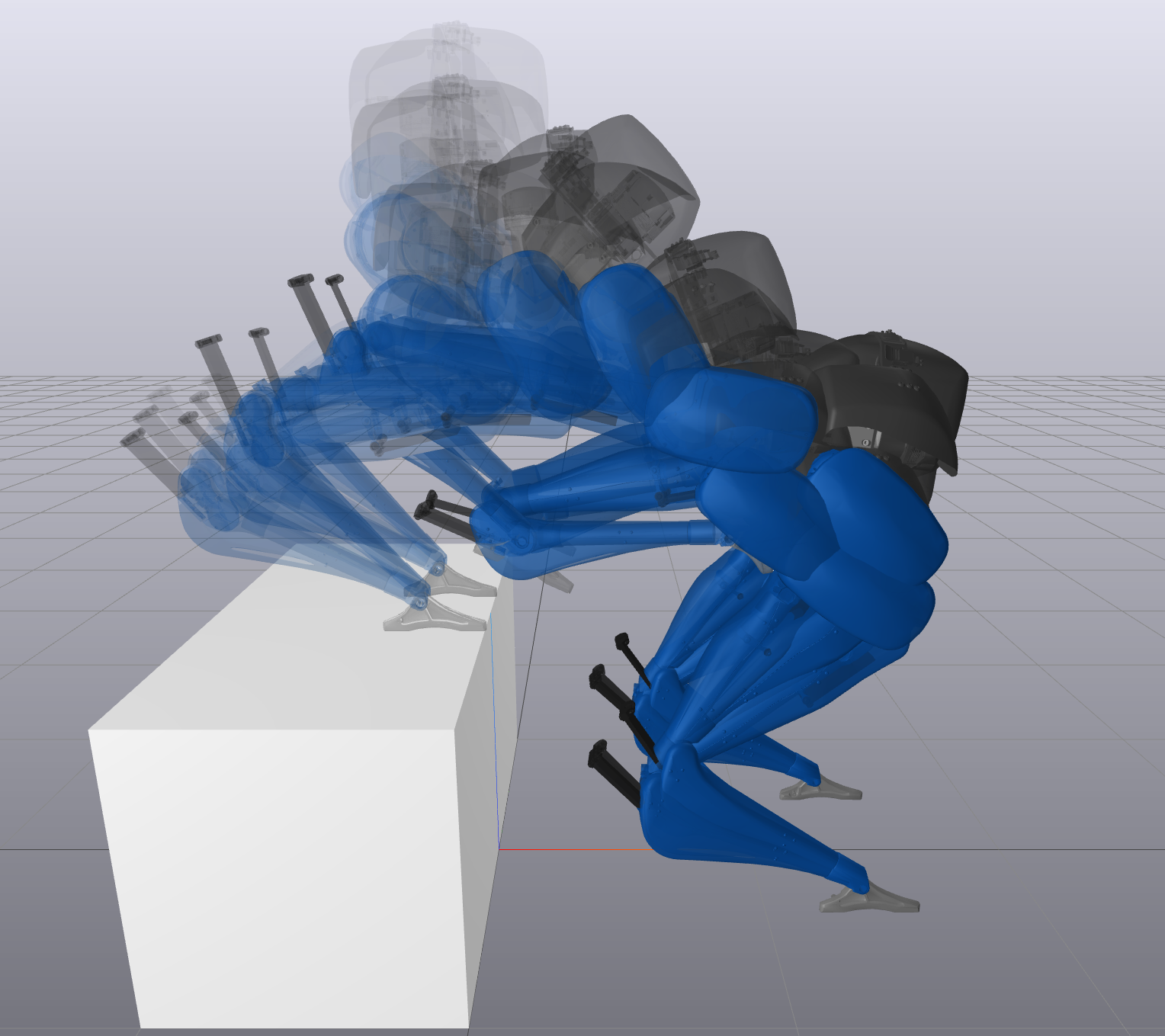}
			\caption{Down jump}
		\end{subfigure}
		\caption{Reference trajectories generated using full model trajectory optimization.}
		\label{fig:reference_jumping_trajectories}
	\end{figure*}
	
	\change{
		To showcase the advantages of using the impact-invariant subspace, we apply the aforementioned projection on the following examples, each organized in the following sections:
		\begin{itemize}
			\item In \cref{sec:five_link_biped}, we consider, in simulation, a joint-space inverse dynamics walking controller for the planar five-link biped Rabbit \cite{chevallereau2003rabbit}, we compare impact-invariant control to a default controller that makes no adjustment near impacts and to a controller that applies no-derivative feedback near impact.
			\item In \cref{sec:jumping_controller}, we formulate an operational space jumping controller for the 3D bipedal robot Cassie. We consider a basic jump as well as more dynamic jumps such as a long jump, box jump, and down jump. 
			In simulation, we evaluate the long jump and down jump controllers through a parameter study.
			On hardware, we validate the jump and box jump controllers with impact-invariant control.
			\item \cref{sec:running_controller}, we develop an operational space running controller for Cassie, which we evaluate in simulation and on hardware.
		\end{itemize}
	}
	
	\subsection{Baseline Comparisons}
	
	\change{
		We include three controllers in our evaluations:
		\begin{itemize}
			\item The \textbf{default controller} is a controller that makes no special considerations with regards to the impact event other than switching contact modes at the nominal impact time. In other words, the standard hybrid controller.
			\item The \textbf{no derivative feedback} controller is a controller that sets all derivative gains to zero ($K_d = 0$) in a window before and after the nominal impact time. For proper comparisons, we use the same window between the no derivative feedback controller and the impact-invariant controller.
			\item The \textbf{impact-invariant projection} controller is our proposed method. We blend in the projection in a window before and after the nominal impact time.
		\end{itemize}
		The \textbf{default} and \textbf{no derivative feedback} controllers serve as the baselines for our comparisons.
		They represent either side of the extremes, where the default controller is extremely sensitive to uncertainty in the impact event but relinquishes no control authority, while the no derivative feedback controller gives up entirely on tracking velocities during impacts but loses significant control authority.
		Although eliminating derivative feedback near impacts may seem extreme, the intermediate mode described in \cite{van2023dual} is exactly this choice of controller.
	}
	
	\subsection{Evaluation Metrics}
	
	\change{
		We evaluate the three controllers (default, no derivative feedback, impact-invariant) on three metrics of decreasing importance: stability, tracking performance, and controller effort.
	}
	
	\change{
		In this paper, we define stability as a binary value of whether or not the robot falls when executing the motion.
		We consider stability to be the primary metric, as falling for legged robots is often considered catastrophic failure.
		Minimizing tracking error is the objective of our controllers as it is essential for stability, so naturally it is our second most important evaluation metric.
		However, a key insight of our method is that there is uncertainty in both our target state and our measured state during impacts, and thus the standard measure of tracking error, particularly in the velocities, may not necessarily be a useful metric.
		For these reasons, we selectively report what we consider are the relevant tracking errors for each experiment.
		Finally, for the jumping experiments, the whole body motion is significantly more sensitive to the initial leap than to the control actions taken during the impact event when the robot lands.
		For this reason, tracking error is a poor metric to compare how the robot performs near the impact event, which is the focus of this paper.
		Instead, we evaluate the controller performance using the control efforts taken near the impact event, considering large controller spikes to be undesirable as those are prone to damaging the robot.
	}
	
	\section{Five-Link Biped Walking Controller}
	\label{sec:five_link_biped}
	
	\change{
		We showcase the benefits of our method on a joint-space inverse dynamics controller for the planar five-link biped.
		This simple system and controller serves as a controlled example to demonstrate the  directional nature of the impact-invariant projection.
	}
	
	\subsection{Experimental Setup}
	\change{
		In simulation, we model our planar five-link biped using the length, mass and, inertia values of Rabbit specified in \cite{chevallereau2003rabbit}, replicated in \cref{tab:rabbit_parameters} for completeness.
		The joint-space controller is tasked with tracking the hip and knee joint angles in both legs for a total of 4 tracking objectives to match the degrees of actuation.
	}
	We generate a periodic walking trajectory using trajectory optimization and perturb the swing foot vertical velocity by 0.1 m/s at the start of the trajectory, which causes the robot to makes contact approximately 5ms after the nominal impact time.
	
	\subsection{Results}
	
	\begin{figure}[h]
		\centering
		\includegraphics[width=0.48\textwidth]{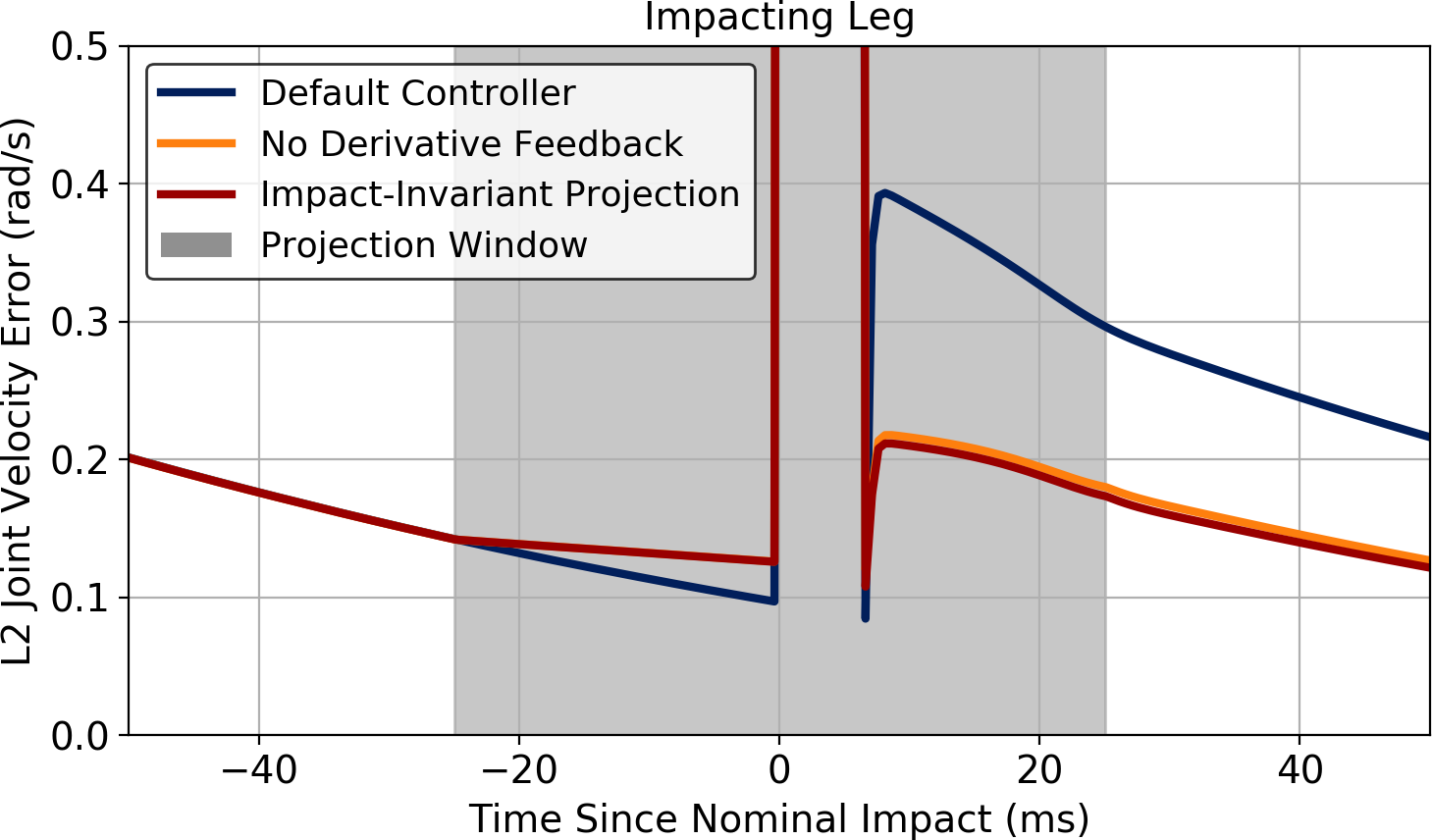}
		\includegraphics[width=0.48\textwidth]{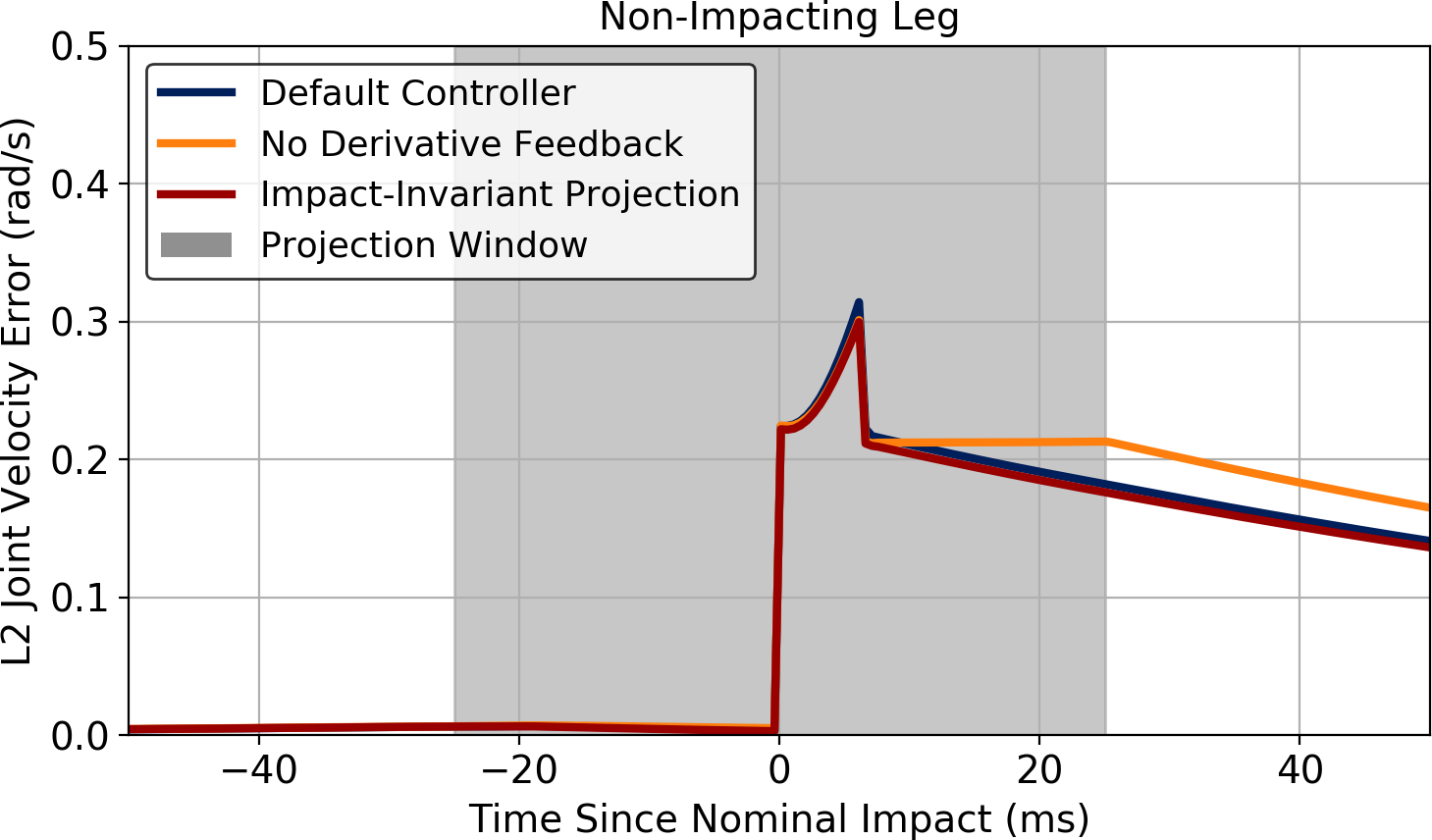}
		\caption{The joint velocity tracking errors are shown for the impacting leg (top) and the non-impacting leg (bottom) for  three strategies. The controller that utilizes the impact-invariant projection is shown to be robust to the mismatch in impact timing as evidenced by lower tracking error compared to the default controller. The impact-invariant controller is also able to maintain full control authority over the joints in the non-impacting leg compared to the controller that applies no derivative feedback in the same window. The time window where no-derivative feedback and the impact-invariant projection are active is shown in grey.}
		\label{fig:directional_robustness}
	\end{figure}

	The tracking errors for the joints in the impacting and non-impacting leg are shown in \cref{fig:directional_robustness}.
	The impact-invariant controller is able to achieve the best tracking performance out of the three controllers for \textit{both} legs.
	It has better tracking performance than the default controller for the joint velocities of the impacting leg.
	At the same time, it has better tracking performance than the controller with no derivative feedback for the joint velocities of the non-impacting leg by appropriately regulating the velocities in those joints.

	\section{Cassie Jumping Controller}
	\label{sec:jumping_controller}
	
	Next, we evaluate the performance of the impact-invariant projection on a jumping controller for Cassie.
	We chose to look at jumping due to the richness of the impact event: the robot cannot accurately estimate its state \cite{jeon2022online} when it is in the air and must make impact with the ground with non-zero velocity.
	
	\subsection{Controller Formulation}
	
	\change{
		We formulate the jumping controller as a trajectory tracking problem, where the target trajectory is generated using the full order model and translated to task-space trajectories which are tracked by an operational space controller.
	}
	
	\subsubsection{Reference Trajectories}
	
	The reference jumping trajectories were computed offline by solving a constrained trajectory optimization problem on the full rigid body model of Cassie, excluding the springs.
	While the springs are a significant component to the dynamics, we found that including them made solver convergence extremely difficult.
	
	The various jumping trajectories are distinguished by the constraints imposed:
	\begin{itemize}
		\item The normal jump trajectory was constrained to have the robot pelvis reach an apex height of 0.15 m above its initial starting height and to have both feet reach 15 cm of clearance above the ground at the apex.
		\item The long jump trajectory has the feet land 0.7 m ahead of their initial position.
		\item The box jump trajectory has the feet land on a box that is 0.5 m tall and 0.3 m in front of the starting configuration.
		\item The down jump trajectory has the feet land on a platform 0.5 m below and 0.3 m in front of the starting configuration.
	\end{itemize}
	
	The trajectory optimization problems were transcribed using DIRCON \cite{posa2016optimization} and solved using a combination of IPOPT \cite{wachter2006implementation} and SNOPT  \cite{gill2005snopt}.
	Traces of the reference trajectories are shown in \cref{fig:reference_jumping_trajectories}.
	The same jumping trajectories were used in both simulation and on the physical robot.
	
	\subsubsection{Finite State Machine}
	We decompose the jumping motion into three states {CROUCH, FLIGHT, LAND}, and switch between states at fixed times as computed from the reference trajectories.
	
	\begin{table}[ht]
		\centering
		\caption{Tracking objectives for the jumping controller. All values are expressed in the world frame.}
		\begin{tabular}{lll}
			\toprule
			\textbf{Symbol}&\textbf{Description}\\ 
			\midrule
			$r \in \Real^3$ & Pelvis position\\
			$\psi \in SO(3)$ & Pelvis orientation \\
			$y_{L}, y_{R} \in \Real^3$ & Foot position relative to pelvis \\
			$\beta_{L}, \beta_{R} \in \Real$ & Hip yaw angle \\
			$\phi_{L}, \phi_{R} \in \Real$ & Toe joint angle\\
			\bottomrule
		\end{tabular}
		\label{tab:jumping_objectives}
	\end{table}
	
	\subsubsection{Tracking Objectives}
\begin{figure}[h]
		\includegraphics[width=0.48\textwidth]{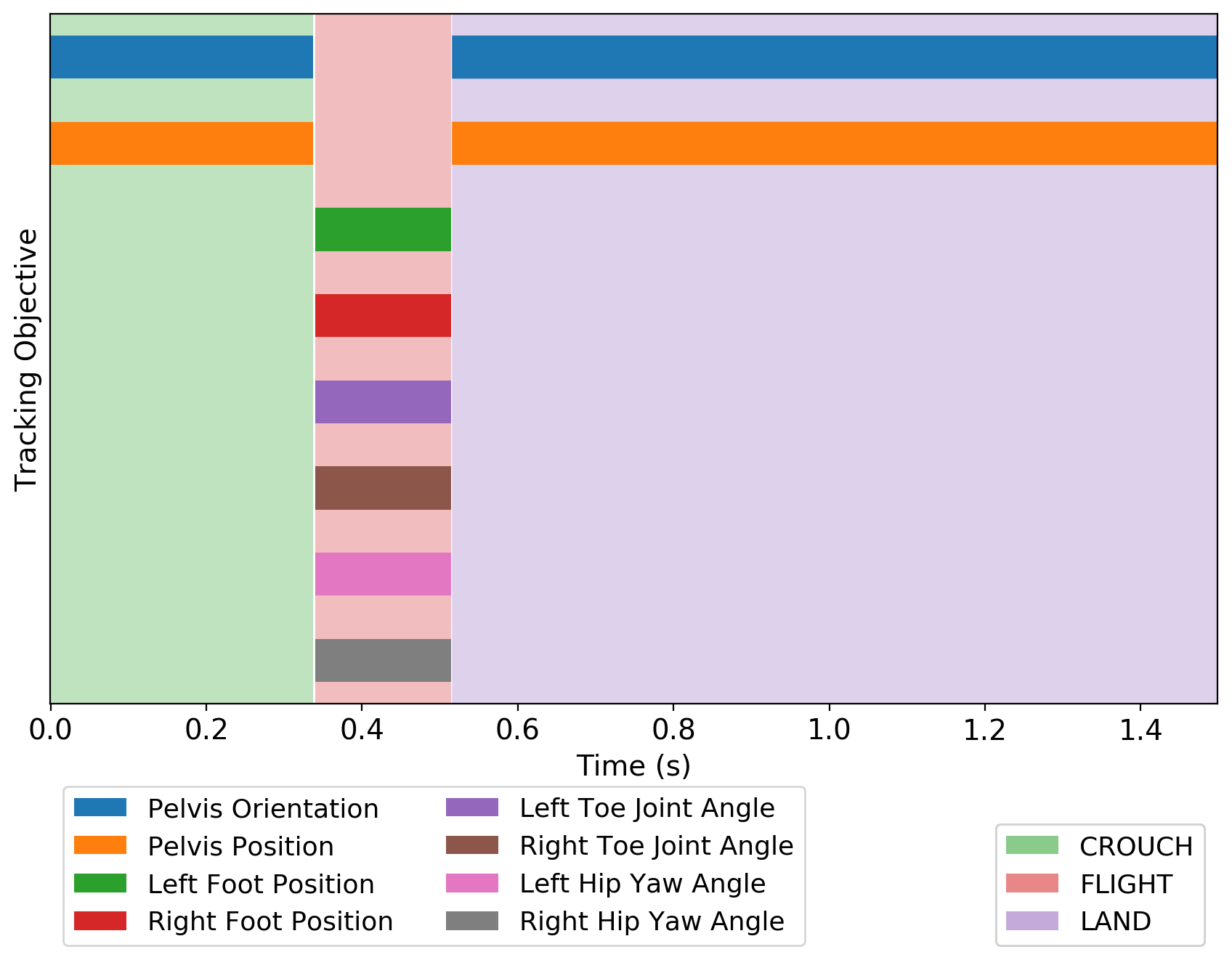}
		\caption{Active tracking objectives per mode for the jumping controller executing the jump trajectory.}
		\label{fig:jumping_active}
	\end{figure}

	The tracking objectives are listed in \cref{tab:jumping_objectives}.
	We define tracking objectives such as the foot and pelvis position relative to other points on the robot to reduce sensitivity to errors in state estimation.
	During the CROUCH and LAND states, the active objectives are \change{$[r, \psi]$.
		During the FLIGHT state, the active objectives are $[y_{L}, y_{R}, \beta_{L}, \beta_{R}, \phi_{L}, \phi_{R}]$}.
	The active tracking objectives per mode are also illustrated in \cref{fig:jumping_active}.
	Similar outputs are used in another jumping controller for Cassie \cite{xiong2018bipedal}.
	
	\subsection{Simulation Results}
	Timing the switch from the FLIGHT state to the LAND state is critical to stabilize the jump. 
	Impact-invariant control reduces sensitivity to the impact timing and thus enables a greater margin of error.
	\change{To evaluate this effect, for all the jumping motions we perturb the transition time $t_{s}$ from the nominal switching time by $[-0.025 s, 0.025 s]$ and evaluate a range of projection duration $T \in [0.0 s, 0.1 s]$ to empirically evaluate the sensitivity of the controller. }
	
\begin{figure}[ht!]
		\begin{subfigure}[b]{0.47\textwidth}
			\includegraphics[width=1.0\textwidth]{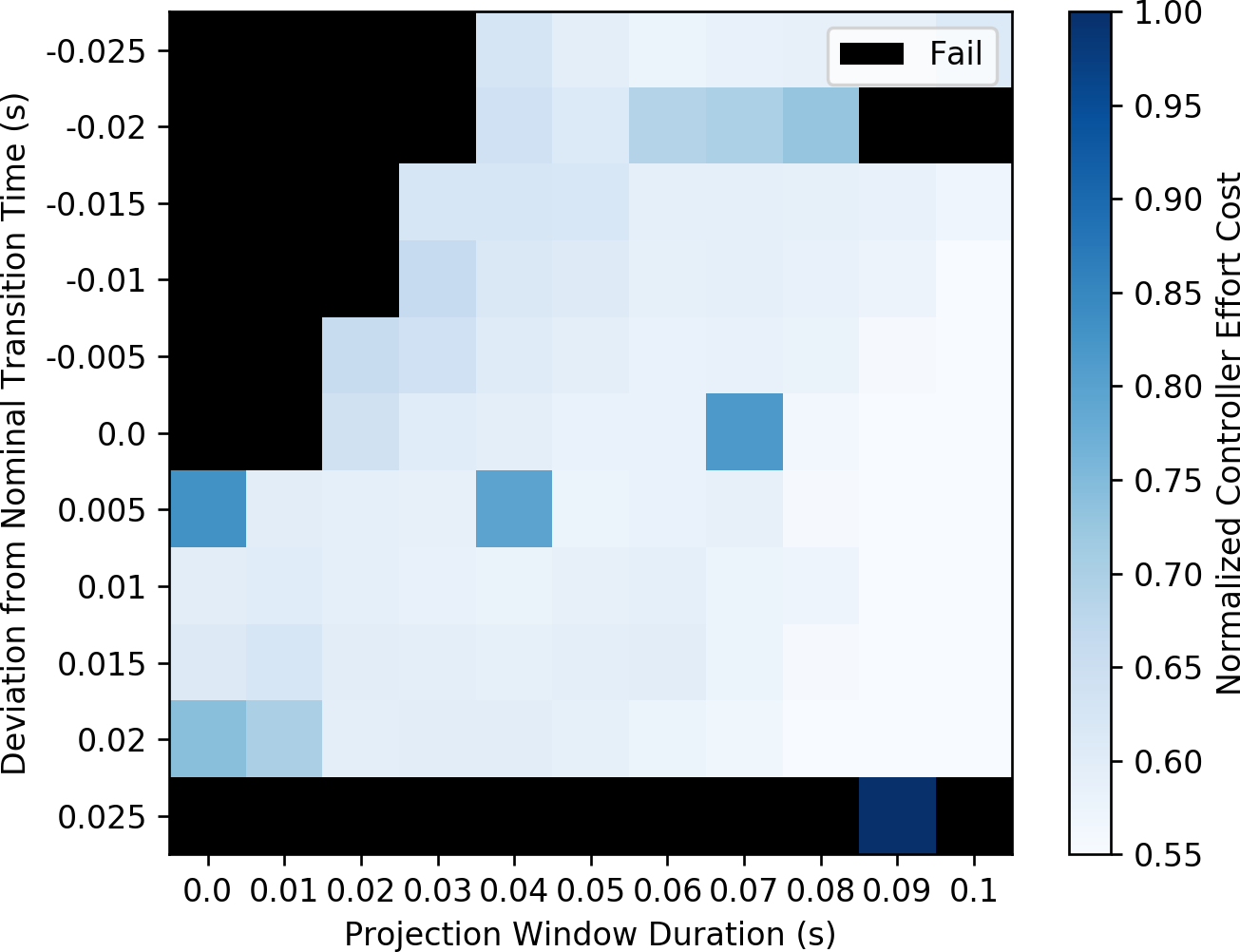}
			\caption{Long jump}
		\end{subfigure}
		\begin{subfigure}[b]{0.47\textwidth}
			\includegraphics[width=1.0\textwidth]{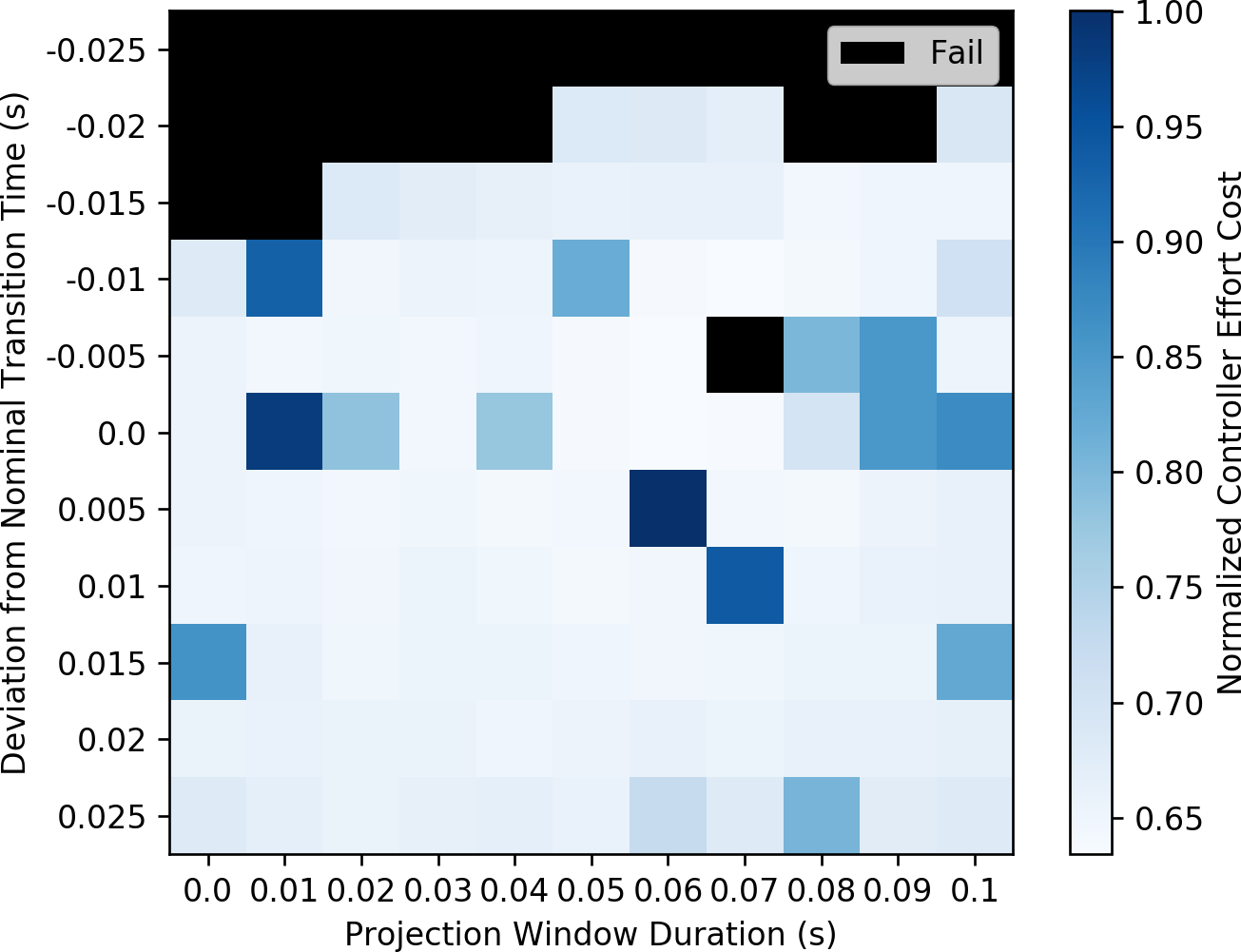}
			\caption{Down jump}
		\end{subfigure}
		\caption{Simulation effort costs for (a) long jump and (b) down jump evaluated over a range of fixed transition times and projection window durations. Five trials are evaluated for each pair of transition times and projection durations. The controller without the impact-invariant projection corresponds to a projection window duration of 0.0 s. The regions where the robot fails to stabilize for over half the trials are marked as ``Fail". As the projection window increases, the range of successful transition times increases. This effect is more pronounced for the long jump.}
		\label{fig:jumping_parameter_study}
	\end{figure}

	In addition to testing whether the robot successfully controls the jump, we also evaluate the overall control input cost:
	\begin{align}
		J_{u} = \int_{t_0}^{t_f} u^T W_u u dt,
	\end{align}
	where $W_u$ is the same regularization weight on $u$ we use in the OSC.
	\change{
		5 experiments per pair of transition times and projection durations are evaluated using the Drake simulator \cite{drake}, and the results are reported in \cref{fig:jumping_parameter_study}, with $J_{u}$ normalized so that the largest value is $1$.
		We exclude cases when there is more than one failure to avoid saturating the cost metric.
	}
	The region of stable jumps increases as the projection window duration increases, while there is not a significant effect on the input cost.
	We see a more noticeable improvement from the impact-invariant controller in the long jump than the down jump. 
	We theorize that this is because the long jump motion is more difficult to stabilize, most likely from less control authority due to friction code limits, and therefore the improved robustness from the impact-invariant controller has a greater marginal effect.
	
	\begin{figure}[h]
		\centering
		\includegraphics[width=0.48\textwidth]{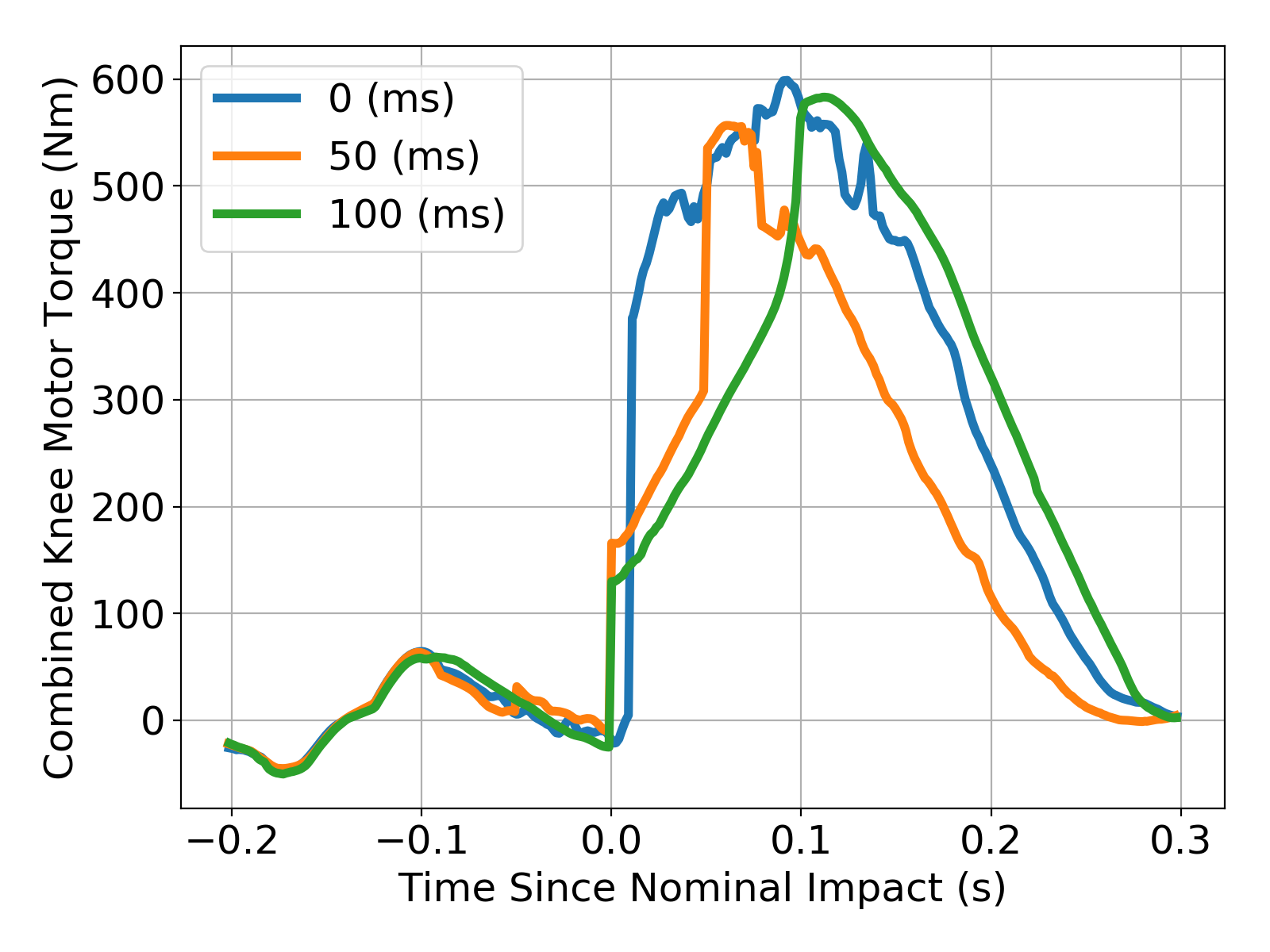}
		\caption{Motor efforts on Cassie executing the default jumping motion. The combined knee motor torques commanded by the jumping controller are shown for three different durations of the impact-invariant projection.}
		\label{fig:hardware_knee_efforts}
	\end{figure}
	
	\subsection{Hardware Results}
	
	Experiments using the controller with and without the impact-invariant projection were both consistently able to successfully complete the basic jump.
	Snapshots of the jumping motion are shown in \cref{fig:cassie_hardware_jump}.
	As seen in \cref{fig:hardware_joint_vels}, the joint velocities change rapidly during the impact event.
	By projecting the velocity error of the outputs (position and orientation of the pelvis) to the impact-invariant subspace, we avoid overreacting to these rapid velocity changes in a principled manner.
	The effects of this can be seen in \cref{fig:hardware_knee_efforts}, where the change in knee motor efforts, particularly at the impact event, are significantly reduced when using the impact-invariant projection.
	We choose to show the knee motor efforts because they exhibit the largest change at the impact event due to their role in absorbing the weight of the pelvis at impact.
	This smoothing out of the commanded motor efforts is similar to what we see in simulation.
	The jumping motions for the controller both with and without the impact-invariant projection are included in the supplementary video.
	
	Among the other jumping trajectories, we chose to test the box jump trajectory on hardware as it is least likely to damage the robot.
	We positioned 16 in ($\sim$0.4 m) tall wooden boxes in front of the robot and executed the same tracking controller using a 50 ms projection duration.
	Although the reference jumping trajectory we optimized was for a box height of 0.5 m, the robustness afforded by the impact-invariant controller enabled the controller to successfully achieve the jump.
	A frame of the controller executing the box jump is shown in \cref{fig:cassie_hardware_jump} and is also included in the supplementary video.
	
	Approximately 20 trials were conducted to tune the controller parameters in order to achieve the first successful jump.
	Due to fear of damaging the robot, we did not conduct enough trials to report a reliability metric for the box jumping controller on hardware.

	\section{Cassie Running Controller}
	\label{sec:running_controller}
	
	Jumping motions have a significant, but singular, impact event. 
	In contrast, running makes impacts with the environment at every stride.
	Because each stride leads immediately into the next, consistent tracking performance is essential for achieving stable running.
	\change{Because bipedal running makes impact with its environment with only a single foot at a time, the space of states that are invariant is also larger for running.
		For these reasons, we develop a running controller as an additional application of the impact-invariant projection.
	}
	
	\subsection{Controller Architecture}
	\label{subsec:running_controller_formulation}
	
\begin{figure}[h]
		\includegraphics[width=0.48\textwidth]{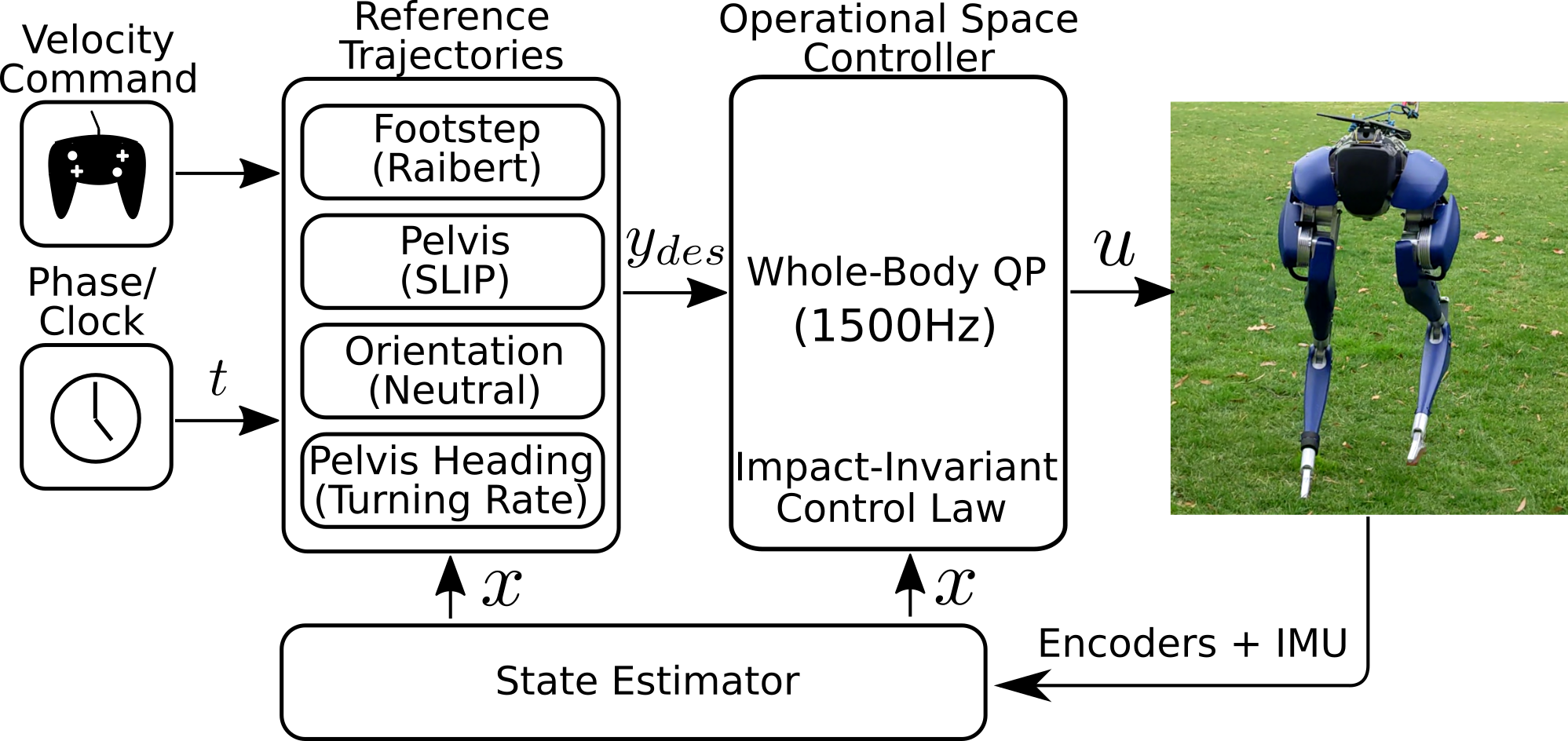}
		\caption{Key elements of the running controller diagram.}
		\label{fig:controller_diagram}
	\end{figure}

	We track a SLIP-like pelvis trajectory and Raibert stepping generated footstep trajectories \cite{raibert1984experiments} using the same OSC framework as used for the Cassie jumping controller.
	A diagram with the key elements is shown in \cref{fig:controller_diagram}.
	
	\subsection{Reference Trajectory Generation}
	\label{subsec:running_ref_traj}
	
	\subsubsection{Mode timing}
	\label{subsubsec:mode_timing}
	We use variable stance and flight durations to construct reference trajectories and determine contact mode switches.
	The nominal stance and flight durations, $T_s$ and $T_f$ respectively, are given in \cref{tab:running_parameters}.
	However, we allow for variations in the timing to better align with the predicted touchdown and liftoff.
	The upcoming transition times for the full gait cycle are computed at each mode switch using the following heuristics:
	
	\change{
		\begin{align}
			T_s^* &= \frac{l}{y_{SLIP}}T_s\\
			T_f^* &= \ddot y_{SLIP} + \sqrt{\dot y_{SLIP}^2 - 2g(l - y_{SLIP})},
		\end{align}
		where we clip the modified stance and flight durations between $(1 \pm \sigma_{s}) T_s$ and $(1 \pm \sigma_{f}) T_f$ respectively to minimize timing changes in response to large disturbances.
	}
	The modified stance duration computes the ratio of the rest length to the SLIP length to roughly scale the liftoff time. The modified flight duration is computed using the time to touchdown assuming a ballistic trajectory.
	
	\subsubsection{SLIP-like pelvis trajectory}
	Inspired by the extensive literature on SLIP, we use a SLIP-like reference for the pelvis motion during stance \{LS, RS\}.
	We achieve a spring-like behavior by regulating the pelvis position relative to the current stance foot to a constant target height $l$ and using the OSC gains to achieve the desired dynamics.
	\begin{align}
		\ddot y_{SLIP, cmd} = K_p (l - y_{SLIP}) + K_d (\dot y_{SLIP}).
	\end{align}
	An important note is that we tune $K_p$ and $K_d$ to achieve the desired dynamics and not to achieve the best tracking.
	We adopt this simpler approach over tracking to true SLIP dynamics because the true dynamics are more difficult to regulate due to additional parameters and being purely an acceleration reference.
	
	\subsubsection{Footstep trajectories}
	
	Regulating the center of mass velocity is achieved through foot placement.
	While there are possible variations for choosing where to place the foot \cite{kajita2003biped} \cite{gong2022zero}, the basic principle behind all the stepping strategies is stepping in the direction that you are falling.
	We choose to regulate the running velocity by planning footsteps with the widely recognized Raibert footstep control law \cite{raibert1984experiments}:
	\change{
		\begin{align}
			y_{ft,2} := \begin{bmatrix}
				y_{ft,x}\\
				y_{ft,y}\\
				y_{ft,z}\\
			\end{bmatrix} =  \begin{bmatrix}
				\frac{v_{x} T_s}{2} + K_{x} (v_{x} - v_{des,x})\\
				\frac{v_{y} T_s}{2} + K_{y} (v_{y} - v_{des,y})\\
				-l\\
			\end{bmatrix},
			\label{eq:raibert}
		\end{align}
	}
	where $K$ are the Raibert stepping gains, $v_{des}$ are the desired velocities as commanded by the operator, and $v$ is the current pelvis velocity computed by the state estimator. The $x$, $y$, and $z$ subscripts in this context denote the sagittal, lateral, and vertical directions in the pelvis frame respectively.
	Finally, to avoid collisions between the legs, we offset the lateral target foot position by a constant distance $c$ in the respective directions.
	$y_{ft, 2}$ then defines the final target footstep location relative to the pelvis.
	
\begin{figure}[h]
		\includegraphics[width=0.48\textwidth]{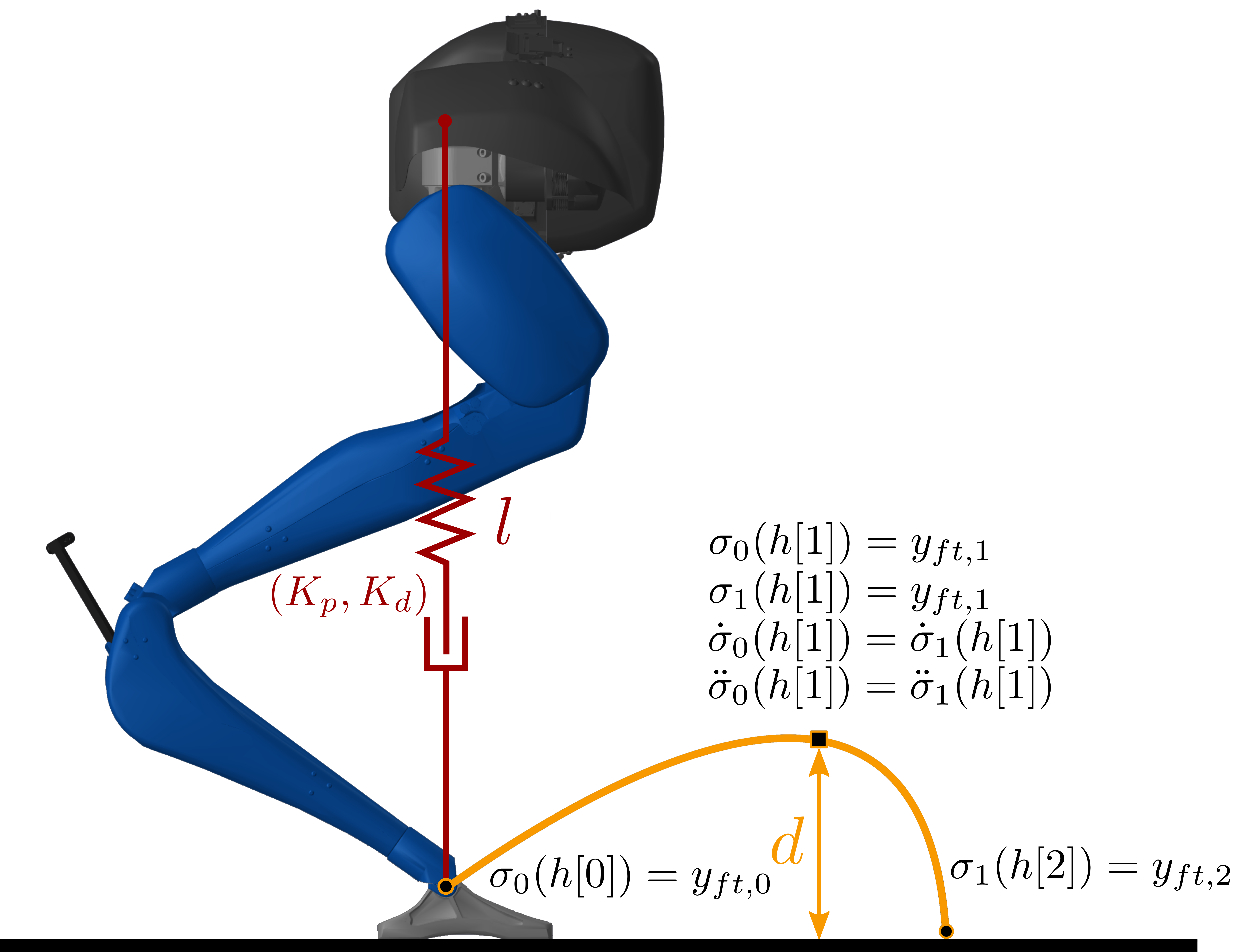}
		\caption{Illustration of key references for the running controller. The target height and swing foot trajectory with leg retraction profiles are shown in red and yellow respectively.}
		\label{fig:running_reference}
	\end{figure}

	With the end footstep location $y_{ft, 2}$ specified, we can generate a trajectory for the swing foot given its initial position $y_{ft, 0}$ at liftoff to the final desired location.
	We specify all the reference trajectories as piecewise quadratic polynomials, so with the additional degrees of freedom we add a waypoint $y_{ft, 1}$ so that the trajectory roughly resembles the swing leg retraction profile observed in both numerical optimization \cite{dai2012optimizing} \cite{seyfarth2003swing} and biology \cite{daley2006running}.
	The full set of constraints defining the piecewise quadratic polynomial are as follows:
	\begin{align*}
		h &= [0, 0.6 (T_s + 2T_f), (T_s + 2T_f)]^T\\
		\sigma_0(h[0]) &= y_{ft, 0}\\
		\sigma_0(h[1]) &= y_{ft, 1}\\
		\sigma_1(h[1]) &= y_{ft, 1}\\
		\sigma_1(h[2]) &= y_{ft, 2}\\
		\dot \sigma_0(h[1]) &= \dot \sigma_1{h[1]}\\
		\ddot \sigma_0(h[1]) &= \ddot \sigma_1{h[1]}
	\end{align*}
	where $y_{ft, 0}$ is the initial foot position at liftoff, $y_{ft, 2}$ is the desired footstep location at touchdown \cref{eq:raibert}, and $y_{ft, 1} = y_{ft, 0} + 0.8 (y_{ft, 0} + y_{ft, 2}) + [0, 0, d]^T$ is the waypoint we introduce to specify foot clearance.
	An illustration of the trajectory profile is shown in \cref{fig:running_reference}. 
	
	\subsection{Reference Tracking}
	
	\subsubsection{Finite State Machine}
	
	\begin{table}[ht]
		\centering
		\caption{Tracking objectives for the running controller. All values are expressed in the world frame.}
		\begin{tabular}{lll}
			\toprule
			\textbf{Symbol}&\textbf{Description}\\ 
			\midrule
			$L_{SLIP} \in \Real^3$ & Pelvis position relative to the stance foot\\
			$\psi \in SO(3)$ & Pelvis Orientation \\
			$y_{L}, y_{R} \in \Real^3$ & Foot position relative to pelvis \\
			$\beta_{L}, \beta_{R} \in \Real$ & Hip yaw angle \\
			$\phi_{L}, \phi_{R} \in \Real$ & Toe joint angle\\
			\bottomrule
		\end{tabular}
		\label{tab:running_objectives}
	\end{table}
	
	We use a time-based finite state machine (FSM) using the timings from \cref{subsubsec:mode_timing} to specify the active contact mode and generate the clock signal for the reference trajectories.
	The set of finite states is \{LS, LF, RS, RF\}; L and R correspond to the left and right legs respectively and S and F correspond to Stance and Flight.
	We distinguish between the two air phases \{LF, RF\} to prescribe different tracking priorities for the different legs.
	The nominal durations for stance and flight deployed on hardware are reported in \cref{tab:running_parameters}.
	
	\begin{table}[h]
		\centering
		\caption{Relevant running controller parameters}
		\begin{tabular}{lll}
			\toprule
			\textbf{Parameter}& \textbf{Symbol} &\textbf{Value}\\ 
			\midrule
			Projection window& $T$ & 0.050 s \\
			Blend time constant& $\tau$ & 0.005 s\\
			Stance duration& $T_s$ & 0.30 s \\
			Flight duration& $T_f$ & 0.09 s \\
			Pelvis target height& $l$ & 0.85 m \\
			Foot clearance& $d$ & 0.2 m \\
			\bottomrule
		\end{tabular}
		\label{tab:running_parameters}
	\end{table}
	
	\subsubsection{Tracking Objectives}
	
	The tracking objectives for the running controller are reported in \cref{tab:running_objectives}.
	Although $L_{SLIP}$ is defined as the position $\in \Real^3$ of the pelvis relative to current stance foot expressed in the world frame, we only track the vertical component.
	
\begin{figure}[ht]
		\includegraphics[width=0.48\textwidth]{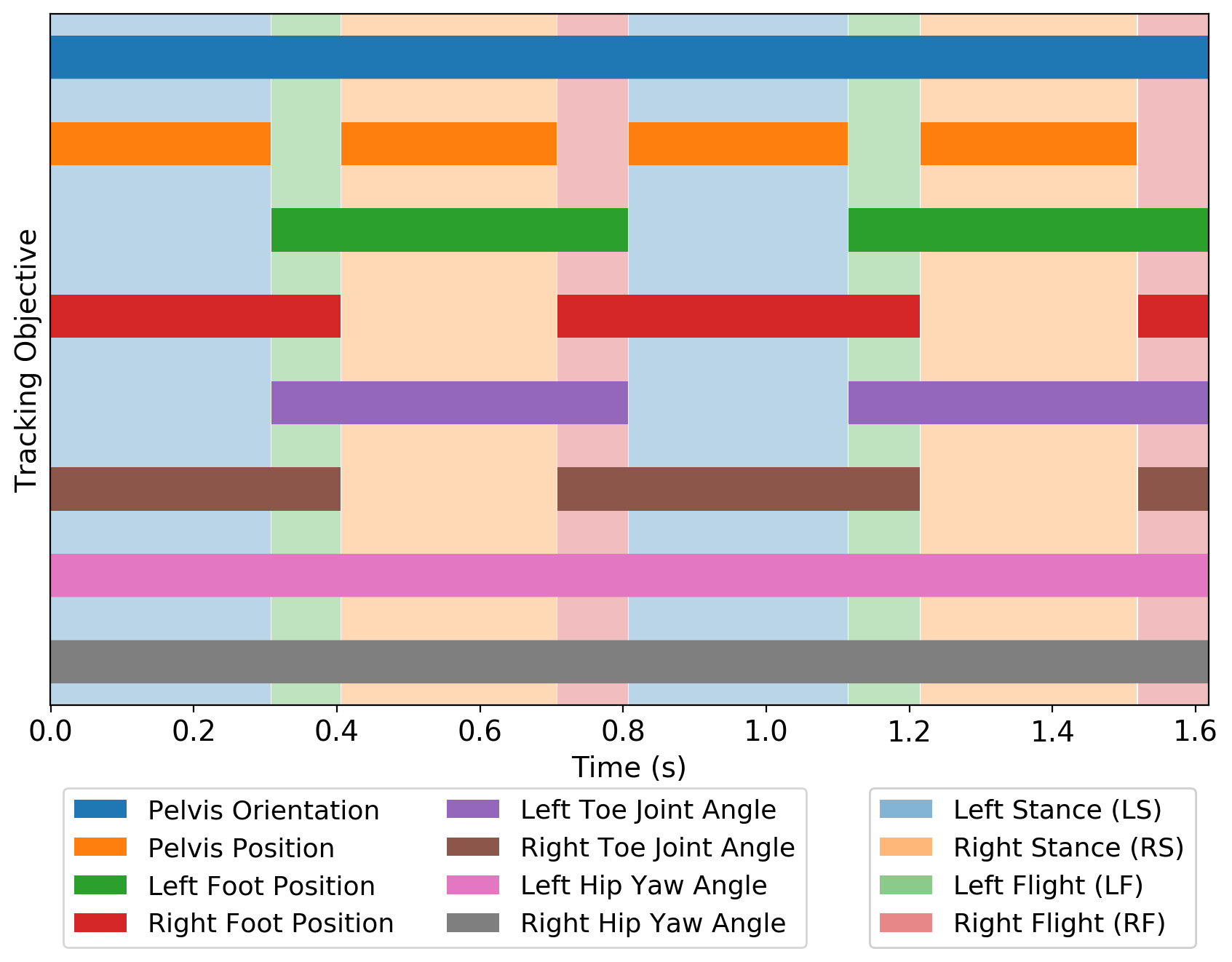}
		\caption{Active tracking objectives per mode for the running controller.}
		\label{fig:running_active}
	\end{figure}

	During left stance (LS), the active vector of tracking objectives is $[L_{SLIP}, y_{R}, \psi, \phi_{R}, \beta_{R}]$.
	For right stance (RS), the tracking objectives are the same, just mirrored for the other leg.
	During the aerial modes {LF, RF}, the active tracking objectives are $[y_{L}, y_{R}, \psi, \beta_{L}, \beta_{R}, \phi_{L}, \phi_{R}]$.
	The active tracking objectives per mode are also illustrated in \cref{fig:running_active}.
	
	Note, the dimension of the tracking objectives in flight, 13, is greater than the total degrees of actuation, 10.
	Therefore, during flight, the control formulation is overspecified and perfect tracking cannot be achieved.
	We choose to leave the problem overspecified as opposed to leaving out either the pelvis orientation or one of the foot targets because we found that trading off between multiple tracking objectives led to better performance on hardware.
	
	\subsubsection{Turning}
	We use the identity quaternion as the desired pelvis orientation and zero as the the desired angular velocity. 
	By setting the task-space gains to $K_p = \text{diag}[150, 200, 0]^T$ and $K_d = \text{diag}[10, 10, 5]^T$, the robot operator can specify the desired yaw velocity and achieve basic turning.
	
	\begin{figure}[ht]
		\includegraphics[width=0.48\textwidth]{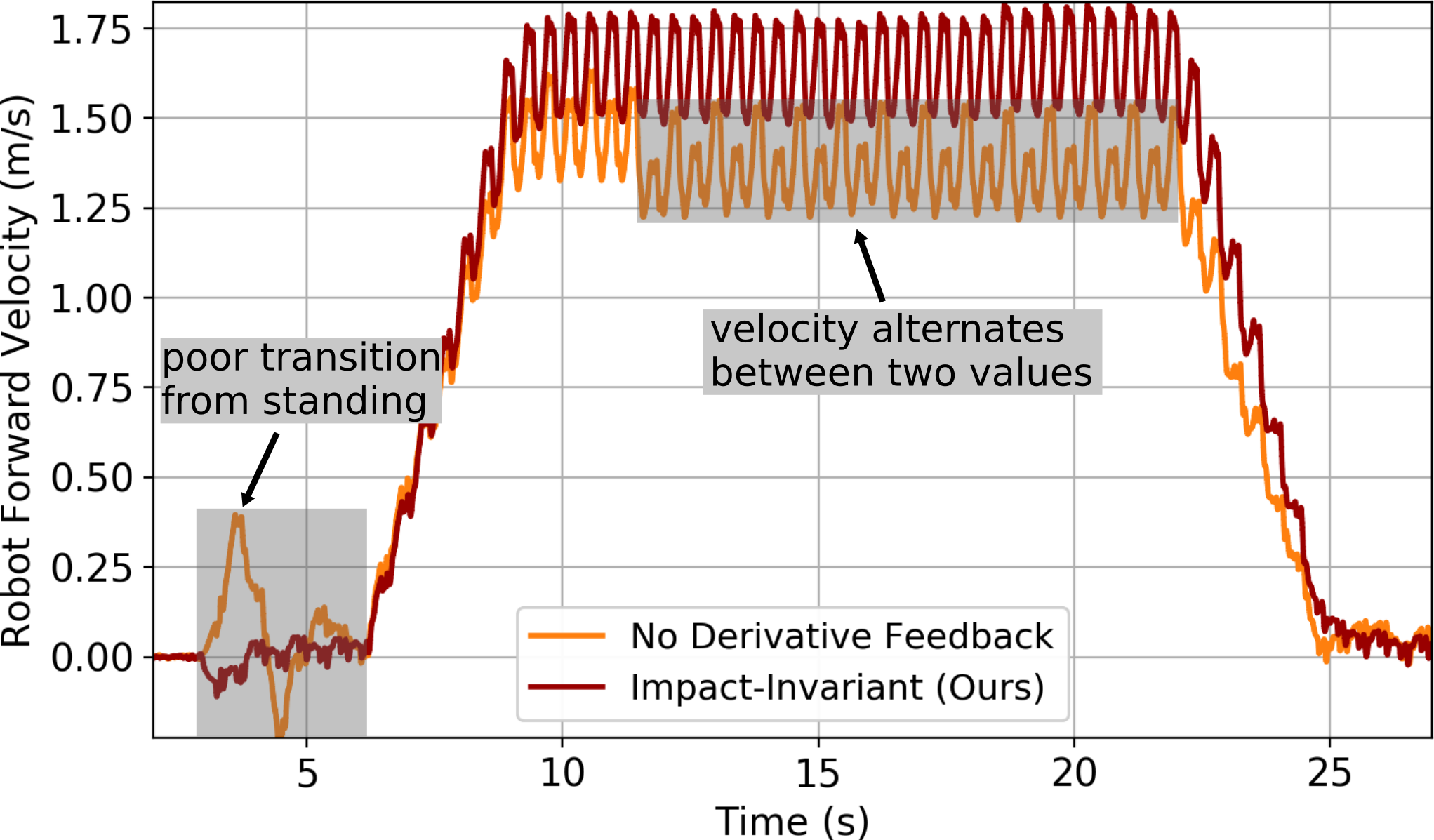}
		\caption{\change{Comparison of our impact-invariant controller to the no-derivative feedback baseline for a simulated robot tracking a target velocity that is linearly ramped up and back down to 0. The velocity profile from the impact-invariant controller is noticeably smoother, which can be directly attributed to better tracking of objectives such as the foot trajectory. At higher velocities, the no-derivative feedback controller alternates between two velocities, which is undesirable. Comparisons to the default controller with no modifications are omitted because to the default controller is entirely unstable.}}
		\label{fig:running_speed}
	\end{figure}
	
	\subsection{Simulation Results}
	
	The projection is an essential component for our framework to achieve stable running.
	\change{Even in simulation, the default controller is immediately unstable when the foot touches the ground due the sensitivity to the impact event being coupled with the severe underactuation\endnote{The running controller operates at the actuation limits of the knee motors and exhibits a significant flight phase, both which exacerbate the underactuation.} of the running gait.
		For this reason, we only report comparisons with the no derivative feedback controller.
	}
	Both controllers are tasked with tracking a forward velocity command and the average errors over 20 secs are reported in \cref{tab:running_comparison}.
	The errors are computed as the position error at touchdown. 
	The positions at touchdown are critical to stable running as they are the primary feedback mechanism for regulating the center of mass.
	\change{
		While the no-derivative controller has only slightly worse tracking performance on individual tracking objectives, the minor tracking discrepancies have a non-negligible effect on overall motion. 
		As shown in \cref{fig:running_speed}, the no derivative controller is around 0.25 m/s slower than the impact-invariant controller with the same velocity command and has noticeably poorer tracking performance.
	}
	
	\begin{table}[ht]
		\centering
		\caption{Average position tracking error compute at each foot touchdown of the impact-invariant controller and no derivative controller for the running controller tracking a constant forward velocity command. Results with the default controller are omitted because the default controller is entirely unstable for the running controller.}
		\resizebox{0.48\textwidth}{!}{
			\begin{tabular}{ccc}
				\toprule
				\textbf{Objective}&\textbf{Impact-Invariant } & \textbf{No Derivative} \\ 
				& \textbf{Error} & \textbf{Error} \\
				\midrule
				Foot forward position & 6.6 cm  & 6.6 cm \\
				Foot lateral position & \textbf{3.6 cm} & 4.4 cm \\
				Foot vertical position & \textbf{3.9 cm} & 5.6 cm \\
				Hip yaw angle & \textbf{0.026 rad} & 0.034 rad \\
				Toe joint angle & 0.030 rad & \textbf{0.020 rad} \\
				\bottomrule
			\end{tabular}
		}
		\label{tab:running_comparison}
	\end{table}

	\subsection{Hardware Results}
	We are able to achieve stable running, with the ability to command forward velocities and turn, on the physical robot using the same gains used in simulation.
	We tested the robot both indoors and outdoors on grass and on a turf field.
	The robustness of the impact-invariant control to early and late impacts on the physical robot is shown in \cref{fig:running_tracking}.
	Videos of the experiments with the physical robot are included in the supplementary video.
	\change{
		The most relevant controller parameters of the running controller are provided in \cref{tab:running_parameters}, and the full running parameters are reported in \cref{sec:appendices} in \cref{tab:full_running_gains}
	}
	
	\begin{figure}[ht]
		\includegraphics[width=0.48\textwidth]{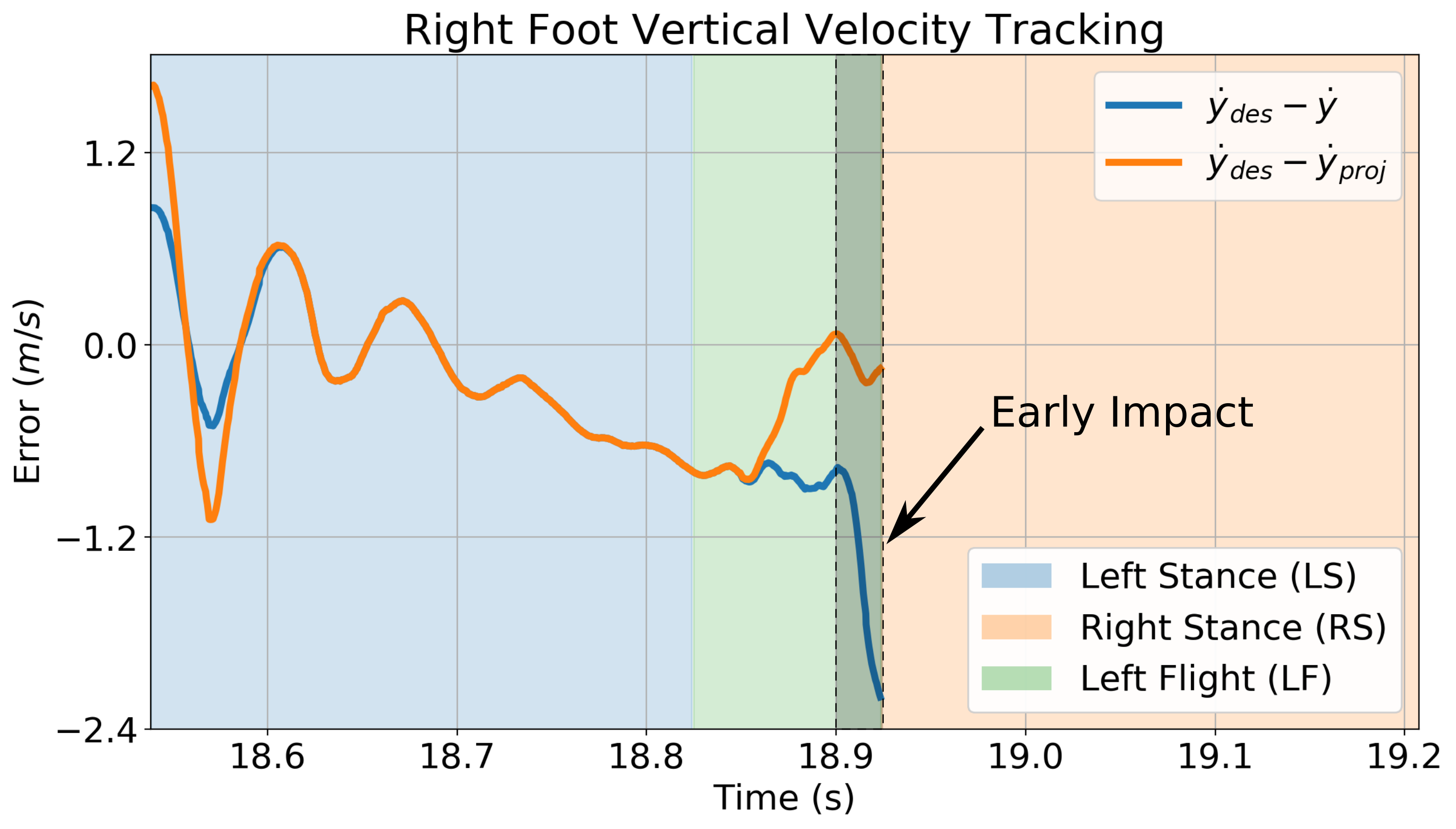}
		\includegraphics[width=0.48\textwidth]{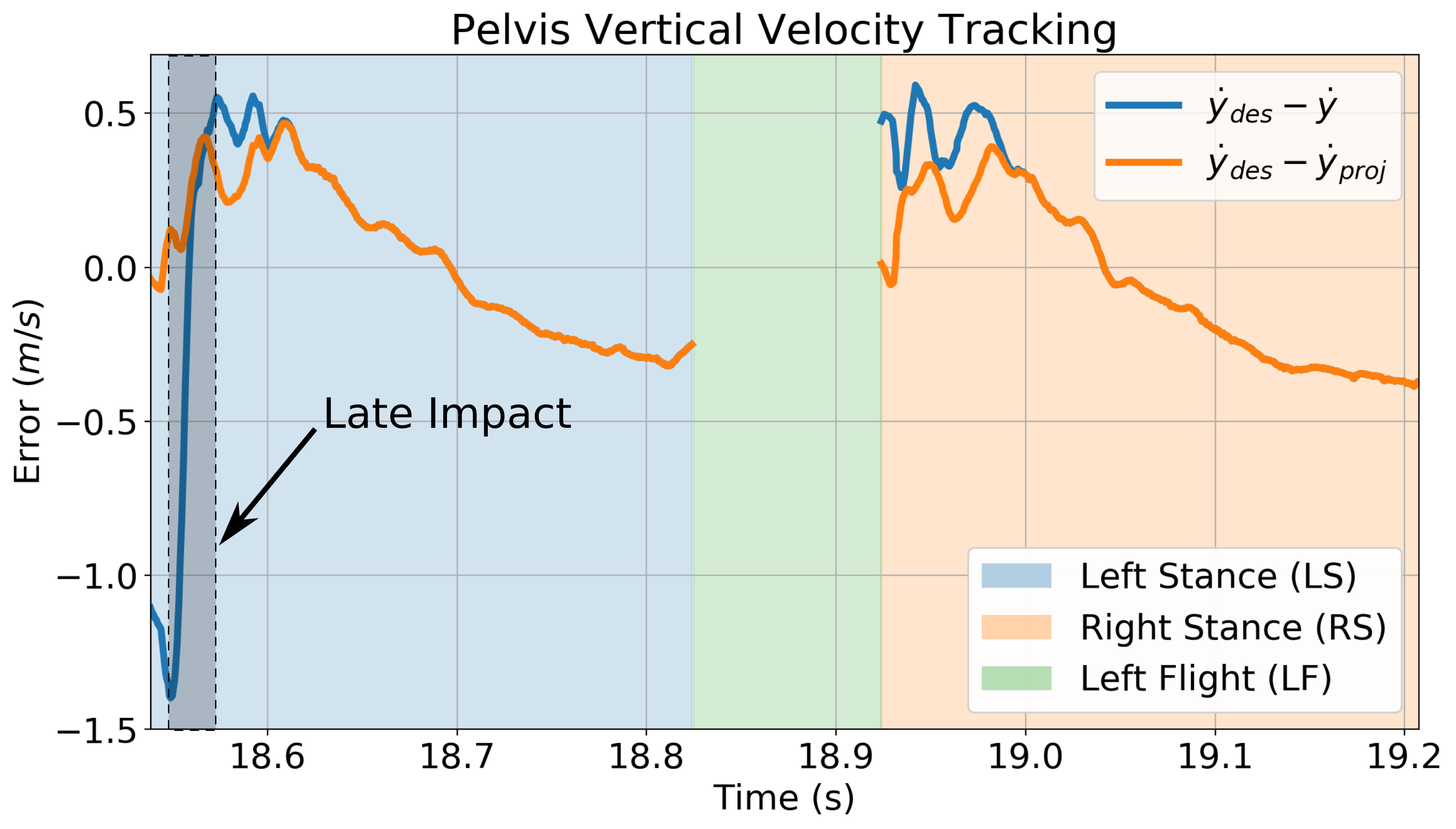}
		\caption{Velocity tracking errors from the running controller on the physical Cassie robot. During the experiment, the robot makes impact \textit{late} with its left foot, which initially results in a large tracking error for the pelvis target. The robot also make impact \textit{early} with its right foot, which results in a large tracking error for the right foot target. In both cases, the impact-invariant projection protects the controller from overreacting to the mismatch in impact timing.}
		\label{fig:running_tracking}
	\end{figure}

	\section{Discussion}
	\label{sec:discussion}
	
	In this paper, we introduce a general strategy that enables controllers to be robust to uncertainties in the impact event without sacrificing control authority over unaffected dimensions.
	The strategy makes use of an easy-to-compute modification of how the velocities of the robot enter the controller.
	Specifically, we project the velocity error into a subspace that is invariant to the impact event.
	This projection completely eliminates sensitivity of the controller to potential contact impulses, while minimally deviating from the original controller.
	
	We demonstrate through examples with legged robots that the impact-invariant projection is robust to the impact event, while achieving better tracking performance compared to alternative methods. The modification can easily be applied to controllers for complex bipeds such as Cassie, and the modification enables highly dynamic motions such as jumping and running.
	
	\subsection{Recovered Control Authority}
	We only apply the projection in a small time window around an impact. 
	An alternative to the projection is to instead turn off all derivative feedback as is done during the contact transition mode \cite{van2022robot}. 
	If impacts are infrequent and short, the additional
	benefit gained from using the impact-invariant projection is minimal. 
	However, for motions where impacts are frequent such as running, the additional benefit can be substantial. 
	Our running controller uses a projection window of 50 ms on both sides of the impact event for a total of 100 ms when
	the projection is active. 
	The nominal stepping period of the running controller is 0.39 s, so the projection is active for over 25\% of the time. 
	Even tighter window durations cover a non-negligible duration of the entire motion.
	
	\subsection{Introducing Additional Controller Invariances}
	
	The impact-invariant framework enables robustness to a very particular source of uncertainty: impacts.
	It seems straightforward to extend this idea to eliminate sources of uncertainty from the control law, but on further examination impacts may be a very specific case of where this method would be useful.
	The key observation of this work is that impacts are \textit{brief} periods of \textit{high uncertainty} that enter the dynamics in a highly structured manner.
	Therefore, a reasonable approach is to effectively ignore the uncertainty in the short amount of time when the magnitude of the uncertainty is at its greatest.
	Other sources of uncertainty such as model differences or uncontrollable elements such as physical springs do not have these attributes.
	We cannot ignore model differences as they permeate the entire dynamics and are persistent throughout, and while springs only enter the dynamics at a single joint, their oscillations do not resolve in a short amount of time.
	
	\subsection{Future Work}
	Although the examples featured in this work focus on legged locomotion, the method is general and can be applied to other rigid body systems with impacts, such as manipulation.
	For future work, we plan to investigate how this method can be applied to high-speed grasping.
	A complementary avenue of future research lies in how we can leverage other sensor modalities such as tactile sensing as a method to have a less conservative bound on impact uncertainty by working with partial sensor feedback.
	
	\change{
		Reinforcement learning has produced controllers with similar dynamic motions on Cassie \cite{siekmann2021sim} \cite{li2024reinforcement} with impressive robustness.
		Another avenue of future research is to evaluate whether these learned controllers exhibit similar invariances and whether incorporating such invariances into the learned controllers might improve robustness.
	}

	\begin{acks}
		We thank Yu-Ming Chen and Brian Acosta for their countless hours of help with hardware testing and helpful discussions. We thank Alp Aydinoglu for helpful discussions about the robust control formulation.
		
		\change{
			This material is based upon work supported by the National Science Foundation Graduate Research Fellowship Program under Grant No. DGE-1845298. Additional funding was provided from Toyota Research Institute.}
	\end{acks}
	
	\theendnotes
	\bibliographystyle{SageH}
	\bibliography{references.bib}

\begin{thebibliography}{47}
\providecommand{\natexlab}[1]{#1}
\providecommand{\url}[1]{\texttt{#1}}
\providecommand{\urlprefix}{URL }
\expandafter\ifx\csname urlstyle\endcsname\relax
  \providecommand{\doi}[1]{DOI:\discretionary{}{}{}#1}\else
  \providecommand{\doi}{DOI:\discretionary{}{}{}\begingroup
  \urlstyle{rm}\Url}\fi

\bibitem[{Acosta et~al.(2022)Acosta, Yang and Posa}]{acosta2022validating}
Acosta B, Yang W and Posa M (2022) Validating robotics simulators on real-world
  impacts.
\newblock \emph{IEEE Robotics and Automation Letters} 7(3): 6471--6478.

\bibitem[{Ames et~al.(2019)Ames, Coogan, Egerstedt, Notomista, Sreenath and
  Tabuada}]{ames2019control}
Ames AD, Coogan S, Egerstedt M, Notomista G, Sreenath K and Tabuada P (2019)
  Control barrier functions: Theory and applications.
\newblock In: \emph{2019 18th European control conference (ECC)}. IEEE, pp.
  3420--3431.

\bibitem[{Atkeson et~al.(2015)Atkeson, Babu, Banerjee, Berenson, Bove, Cui,
  DeDonato, Du, Feng, Franklin, Gennert, Graff, He, Jaeger, Kim, Knoedler, Li,
  Liu, Long, Padir, Polido, Tighe and Xinjilefu}]{atkeson2015no}
Atkeson CG, Babu BPW, Banerjee N, Berenson D, Bove CP, Cui X, DeDonato M, Du R,
  Feng S, Franklin P, Gennert M, Graff JP, He P, Jaeger A, Kim J, Knoedler K,
  Li L, Liu C, Long X, Padir T, Polido F, Tighe GG and Xinjilefu X (2015) No
  falls, no resets: Reliable humanoid behavior in the darpa robotics challenge.
\newblock In: \emph{2015 IEEE-RAS 15th International Conference on Humanoid
  Robots (Humanoids)}. pp. 623--630.
\newblock \doi{10.1109/HUMANOIDS.2015.7363436}.

\bibitem[{Bledt et~al.(2018{\natexlab{a}})Bledt, Powell, Katz, Di~Carlo,
  Wensing and Kim}]{bledt2018cheetah}
Bledt G, Powell MJ, Katz B, Di~Carlo J, Wensing PM and Kim S
  (2018{\natexlab{a}}) Mit cheetah 3: Design and control of a robust, dynamic
  quadruped robot.
\newblock In: \emph{2018 IEEE/RSJ International Conference on Intelligent
  Robots and Systems (IROS)}. IEEE, pp. 2245--2252.

\bibitem[{Bledt et~al.(2018{\natexlab{b}})Bledt, Wensing, Ingersoll and
  Kim}]{bledt2018contact}
Bledt G, Wensing PM, Ingersoll S and Kim S (2018{\natexlab{b}}) Contact model
  fusion for event-based locomotion in unstructured terrains.
\newblock In: \emph{2018 IEEE International Conference on Robotics and
  Automation (ICRA)}. IEEE, pp. 1--8.

\bibitem[{Boyd and Vandenberghe(2004)}]{boyd2004convex}
Boyd SP and Vandenberghe L (2004) \emph{Convex optimization}.
\newblock Cambridge university press.

\bibitem[{Chevallereau et~al.(2003)Chevallereau, Abba, Aoustin, Plestan,
  Westervelt, De~Wit and Grizzle}]{chevallereau2003rabbit}
Chevallereau C, Abba G, Aoustin Y, Plestan F, Westervelt E, De~Wit CC and
  Grizzle J (2003) Rabbit: A testbed for advanced control theory.
\newblock \emph{IEEE Control Systems Magazine} 23(5): 57--79.

\bibitem[{Dai and Tedrake(2012)}]{dai2012optimizing}
Dai H and Tedrake R (2012) Optimizing robust limit cycles for legged locomotion
  on unknown terrain.
\newblock In: \emph{2012 IEEE 51st IEEE Conference on Decision and Control
  (CDC)}. IEEE, pp. 1207--1213.

\bibitem[{Daley and Biewener(2006)}]{daley2006running}
Daley MA and Biewener AA (2006) Running over rough terrain reveals limb control
  for intrinsic stability.
\newblock \emph{Proceedings of the National Academy of Sciences} 103(42):
  15681--15686.

\bibitem[{Fazeli et~al.(2020)Fazeli, Zapolsky, Drumwright and
  Rodriguez}]{fazeli2020fundamental}
Fazeli N, Zapolsky S, Drumwright E and Rodriguez A (2020) Fundamental
  limitations in performance and interpretability of common planar rigid-body
  contact models.
\newblock In: \emph{International Symposium on Robotics Research (ISRR)}.
  Springer, pp. 555--571.

\bibitem[{Featherstone(2014)}]{featherstone2014rigid}
Featherstone R (2014) \emph{Rigid body dynamics algorithms}.
\newblock Springer.
\newblock
  \urlprefix\url{https://link.springer.com/book/10.1007/978-1-4899-7560-7}.

\bibitem[{Gill et~al.(2005)Gill, Murray and Saunders}]{gill2005snopt}
Gill PE, Murray W and Saunders MA (2005) Snopt: An sqp algorithm for
  large-scale constrained optimization.
\newblock \emph{SIAM review} 47(1): 99--131.

\bibitem[{Gong and Grizzle(2022)}]{gong2022zero}
Gong Y and Grizzle JW (2022) Zero dynamics, pendulum models, and angular
  momentum in feedback control of bipedal locomotion.
\newblock \emph{Journal of Dynamic Systems, Measurement, and Control} 144(12):
  121006.

\bibitem[{Green et~al.(2020)Green, Hatton and Hurst}]{green2020planning}
Green K, Hatton RL and Hurst J (2020) Planning for the unexpected: Explicitly
  optimizing motions for ground uncertainty in running.
\newblock In: \emph{2020 IEEE International Conference on Robotics and
  Automation (ICRA)}. IEEE, pp. 1445--1451.

\bibitem[{Halm and Posa(2019)}]{halm2019modeling}
Halm M and Posa M (2019) Modeling and analysis of non-unique behaviors in
  multiple frictional impacts.
\newblock In: \emph{Robotics: Science and Systems}.

\bibitem[{Hartley et~al.(2020)Hartley, Ghaffari, Eustice and
  Grizzle}]{hartley2020contact}
Hartley R, Ghaffari M, Eustice RM and Grizzle JW (2020) Contact-aided invariant
  extended kalman filtering for robot state estimation.
\newblock \emph{The International Journal of Robotics Research} 39(4):
  402--430.

\bibitem[{Huang et~al.(2010)Huang, Olson and Moore}]{huang2010lcm}
Huang AS, Olson E and Moore DC (2010) Lcm: Lightweight communications and
  marshalling.
\newblock In: \emph{2010 IEEE/RSJ International Conference on Intelligent
  Robots and Systems}. IEEE, pp. 4057--4062.

\bibitem[{Jeon et~al.(2022)Jeon, Kim and Kim}]{jeon2022online}
Jeon SH, Kim S and Kim D (2022) Online optimal landing control of the mit mini
  cheetah.
\newblock In: \emph{2022 International Conference on Robotics and Automation
  (ICRA)}. IEEE, pp. 178--184.

\bibitem[{Kajita et~al.(2003)Kajita, Kanehiro, Kaneko, Fujiwara, Harada, Yokoi
  and Hirukawa}]{kajita2003biped}
Kajita S, Kanehiro F, Kaneko K, Fujiwara K, Harada K, Yokoi K and Hirukawa H
  (2003) Biped walking pattern generation by using preview control of
  zero-moment point.
\newblock In: \emph{2003 IEEE International Conference on Robotics and
  Automation (Cat. No. 03CH37422)}, volume~2. IEEE, pp. 1620--1626.

\bibitem[{Kajita et~al.(2001)Kajita, Kanehiro, Kaneko, Yokoi and
  Hirukawa}]{kajita20013d}
Kajita S, Kanehiro F, Kaneko K, Yokoi K and Hirukawa H (2001) The 3d linear
  inverted pendulum mode: A simple modeling for a biped walking pattern
  generation.
\newblock In: \emph{Proceedings 2001 IEEE/RSJ International Conference on
  Intelligent Robots and Systems. Expanding the Societal Role of Robotics in
  the the Next Millennium (Cat. No. 01CH37180)}, volume~1. IEEE, pp. 239--246.

\bibitem[{Kong et~al.(2023)Kong, Payne, Zhu and Johnson}]{kong2023saltation}
Kong NJ, Payne JJ, Zhu J and Johnson AM (2023) Saltation matrices: The
  essential tool for linearizing hybrid dynamical systems.
\newblock \emph{arXiv preprint arXiv:2306.06862} .

\bibitem[{Li and Wensing(2020)}]{li2020hybrid}
Li H and Wensing PM (2020) Hybrid systems differential dynamic programming for
  whole-body motion planning of legged robots.
\newblock \emph{IEEE Robotics and Automation Letters} 5(4): 5448--5455.

\bibitem[{Li et~al.(2024)Li, Peng, Abbeel, Levine, Berseth and
  Sreenath}]{li2024reinforcement}
Li Z, Peng XB, Abbeel P, Levine S, Berseth G and Sreenath K (2024)
  Reinforcement learning for versatile, dynamic, and robust bipedal locomotion
  control.
\newblock \emph{arXiv preprint arXiv:2401.16889} .

\bibitem[{Mason et~al.(2016)Mason, Rotella, Schaal and
  Righetti}]{mason2016balancing}
Mason S, Rotella N, Schaal S and Righetti L (2016) Balancing and walking using
  full dynamics lqr control with contact constraints.
\newblock In: \emph{2016 IEEE-RAS 16th International Conference on Humanoid
  Robots (Humanoids)}. IEEE, pp. 63--68.

\bibitem[{Posa et~al.(2016)Posa, Kuindersma and Tedrake}]{posa2016optimization}
Posa M, Kuindersma S and Tedrake R (2016) Optimization and stabilization of
  trajectories for constrained dynamical systems.
\newblock In: \emph{2016 IEEE International Conference on Robotics and
  Automation (ICRA)}. IEEE, pp. 1366--1373.

\bibitem[{Raibert et~al.(1984)Raibert, Brown~Jr and
  Chepponis}]{raibert1984experiments}
Raibert MH, Brown~Jr HB and Chepponis M (1984) Experiments in balance with a 3d
  one-legged hopping machine.
\newblock \emph{The International Journal of Robotics Research} 3(2): 75--92.

\bibitem[{Reher and Ames(2021)}]{reher2021inverse}
Reher J and Ames AD (2021) Inverse dynamics control of compliant hybrid zero
  dynamic walking.
\newblock In: \emph{2021 IEEE International Conference on Robotics and
  Automation (ICRA)}. IEEE, pp. 2040--2047.

\bibitem[{Remy(2017)}]{remy2017ambiguous}
Remy CD (2017) Ambiguous collision outcomes and sliding with infinite friction
  in models of legged systems.
\newblock \emph{The International Journal of Robotics Research} 36(12):
  1252--1267.

\bibitem[{Rijnen et~al.(2017)Rijnen, de~Mooij, Traversaro, Nori, van~de Wouw,
  Saccon and Nijmeijer}]{rijnen2017control}
Rijnen M, de~Mooij E, Traversaro S, Nori F, van~de Wouw N, Saccon A and
  Nijmeijer H (2017) Control of humanoid robot motions with impacts: Numerical
  experiments with reference spreading control.
\newblock In: \emph{2017 IEEE International Conference on Robotics and
  Automation (ICRA)}. IEEE, pp. 4102--4107.

\bibitem[{Salmon(1968)}]{salmon1968}
Salmon D (1968) Minimax controller design.
\newblock \emph{IEEE Transactions on Automatic Control} 13(4): 369--376.
\newblock \doi{10.1109/TAC.1968.1098941}.

\bibitem[{Schwind(1998)}]{schwind1998spring}
Schwind WJ (1998) \emph{Spring loaded inverted pendulum running: A plant
  model}.
\newblock University of Michigan.

\bibitem[{Sentis and Khatib(2005)}]{sentis2005control}
Sentis L and Khatib O (2005) Control of free-floating humanoid robots through
  task prioritization.
\newblock In: \emph{Proceedings of the 2005 IEEE International Conference on
  Robotics and Automation}. IEEE, pp. 1718--1723.

\bibitem[{Seyfarth et~al.(2003)Seyfarth, Geyer and Herr}]{seyfarth2003swing}
Seyfarth A, Geyer H and Herr H (2003) Swing-leg retraction: a simple control
  model for stable running.
\newblock \emph{Journal of Experimental Biology} 206(15): 2547--2555.

\bibitem[{Siekmann et~al.(2021)Siekmann, Godse, Fern and
  Hurst}]{siekmann2021sim}
Siekmann J, Godse Y, Fern A and Hurst J (2021) Sim-to-real learning of all
  common bipedal gaits via periodic reward composition.
\newblock In: \emph{2021 IEEE International Conference on Robotics and
  Automation (ICRA)}. IEEE, pp. 7309--7315.

\bibitem[{Stellato et~al.(2020)Stellato, Banjac, Goulart, Bemporad and
  Boyd}]{stellato2020osqp}
Stellato B, Banjac G, Goulart P, Bemporad A and Boyd S (2020) Osqp: An operator
  splitting solver for quadratic programs.
\newblock \emph{Mathematical Programming Computation} 12(4): 637--672.

\bibitem[{Strang(2022)}]{strang1993introduction}
Strang G (2022) \emph{Introduction to linear algebra}, volume~3.
\newblock SIAM.

\bibitem[{Tedrake and the Drake Development~Team(2019)}]{drake}
Tedrake R and the Drake Development~Team (2019) Drake: Model-based design and
  verification for robotics.
\newblock \urlprefix\url{https://drake.mit.edu}.

\bibitem[{van Steen et~al.(2023)van Steen, Co{\c{s}}gun, van~de Wouw and
  Saccon}]{van2023dual}
van Steen JJ, Co{\c{s}}gun A, van~de Wouw N and Saccon A (2023) Dual arm
  impact-aware grasping through time-invariant reference spreading control.
\newblock \emph{IFAC-PapersOnLine} 56(2): 1009--1016.

\bibitem[{van Steen et~al.(2022)van Steen, van~de Wouw and
  Saccon}]{van2022robot}
van Steen JJ, van~de Wouw N and Saccon A (2022) Robot control for simultaneous
  impact tasks via quadratic programming-based reference spreading.
\newblock In: \emph{2022 American Control Conference (ACC)}. IEEE, pp.
  3865--3872.

\bibitem[{W{\"a}chter and Biegler(2006)}]{wachter2006implementation}
W{\"a}chter A and Biegler LT (2006) On the implementation of an interior-point
  filter line-search algorithm for large-scale nonlinear programming.
\newblock \emph{Mathematical programming} 106: 25--57.

\bibitem[{Wang et~al.(2022)Wang, Dehio and Kheddar}]{wang2022predicting}
Wang Y, Dehio N and Kheddar A (2022) Predicting impact-induced joint velocity
  jumps on kinematic-controlled manipulator.
\newblock \emph{IEEE Robotics and Automation Letters} 7(3): 6226--6233.

\bibitem[{Wang and Kheddar(2019)}]{wang2019impact}
Wang Y and Kheddar A (2019) Impact-friendly robust control design with
  task-space quadratic optimization.
\newblock In: \emph{Robotics: Science and Systems (RSS)}.

\bibitem[{Wensing and Orin(2013)}]{wensing2013generation}
Wensing PM and Orin DE (2013) Generation of dynamic humanoid behaviors through
  task-space control with conic optimization.
\newblock In: \emph{2013 IEEE International Conference on Robotics and
  Automation}. IEEE, pp. 3103--3109.

\bibitem[{Wensing et~al.(2023)Wensing, Posa, Hu, Escande, Mansard and
  Del~Prete}]{wensing2023optimization}
Wensing PM, Posa M, Hu Y, Escande A, Mansard N and Del~Prete A (2023)
  Optimization-based control for dynamic legged robots.
\newblock \emph{IEEE Transactions on Robotics} .

\bibitem[{Xiong and Ames(2018)}]{xiong2018bipedal}
Xiong X and Ames AD (2018) Bipedal hopping: Reduced-order model embedding via
  optimization-based control.
\newblock In: \emph{2018 IEEE/RSJ International Conference on Intelligent
  Robots and Systems (IROS)}. IEEE, pp. 3821--3828.

\bibitem[{Yang and Posa(2021)}]{yang2021impact}
Yang W and Posa M (2021) Impact invariant control with applications to bipedal
  locomotion.
\newblock In: \emph{2021 IEEE/RSJ International Conference on Intelligent
  Robots and Systems (IROS)}. IEEE, pp. 5151--5158.

\bibitem[{Zhu et~al.(2022)Zhu, Kong, Council and Johnson}]{zhu2022hybrid}
Zhu J, Kong NJ, Council G and Johnson AM (2022) Hybrid event shaping to
  stabilize periodic hybrid orbits.
\newblock In: \emph{2022 International Conference on Robotics and Automation
  (ICRA)}. IEEE, pp. 01--07.

\end{thebibliography}
	
	\clearpage
	\appendix
	\section{Appendices}
	\label{sec:appendices}
	
	\begin{table}[h]
		\centering
		\caption{\change{Length, mass and inertia parameters for the planar five-link biped evaluated in simulation. Values are taken from the physical robot, Rabbit \cite{chevallereau2003rabbit}.}}
		\resizebox{0.48\textwidth}{!}{
			\begin{tabular}{llll}
				\toprule
				& thigh & shin & torso\\ 
				\midrule
				length ($m$) & 	0.4 & 0.4 & 0.625\\
				mass ($kg$) & 	6.8 & 3.2 & 12.0\\
				inertia along principal axes ($kg m^2$) & 0.47 & 0.20 & 1.33\\
				\midrule
			\end{tabular}
		}
		\label{tab:rabbit_parameters}
	\end{table}
	
	\begin{table}[h]
		\centering
		\caption{\change{Full impact-invariant parameters and regularization weights for the Cassie jumping controller  }}
		\resizebox{0.48\textwidth}{!}{
			\begin{tabular}{lll}
				\toprule
				\textbf{Symbol}&\textbf{Description} & \textbf{Value}\\ 
				\midrule
				$\mu$ & Friction coefficient & 0.6\\
				$T$ & Projection window & 0.05 s\\
				$\tau$ & Blend time constant & 0.005 s\\
				$W_u$ & Regularization weight on inputs & 1e-6\\
				$W_{acc}$ & Regularization weight on generalized accelerations $\ddot q$ & 1e-4\\
				$W_{\lambda}$ & Regularization weight on contact forces & 1e-1\\
				\midrule
			\end{tabular}
		}
	\end{table}
	\begin{table}[h]
		\centering
		\caption{\change{Feedback gains for the Cassie jumping controller (jump and box jump) deployed on hardware. Weight and gain matrices are diagonal matrices, represented here as vectors to be concise.}}
		\resizebox{0.48\textwidth}{!}{
			\begin{tabular}{llll}
				\toprule
				\textbf{OSC Objective} & $W$ & $K_p$ & $K_d$ \\
				\midrule
				Toe joint angle &  0.01 & 1500 & 10\\
				Hip yaw angle & 2.5 & 100 & 5\\
				Pelvis [x, y, z] & $[20, 2, 20]$ & $[40, 50, 40]$ & $[7.5, 5, 5]$\\
				Pelvis [roll, pitch, yaw] & $[10, 5, 1]$ & $[150, 200, 150]$ & $[10, 10, 5]$\\
				Foot [x, y, z] & $[10, 100, 10]$ & $[125, 50, 150]$ & $[2.5, 2.5, 0]$\\
				\bottomrule
			\end{tabular}
		}
		\label{tab:jumping_gains}
	\end{table}
	
	\begin{table}[h]
		\centering
		\caption{\change{Full trajectory, impact-invariant parameters, and regularization weights for the Cassie running controller deployed on hardware.}}
		\resizebox{0.48\textwidth}{!}{
			\begin{tabular}{lll}
				\toprule
				\textbf{Symbol}&\textbf{Description} & \textbf{Value}\\ 
				\midrule
				$\mu$ & Friction coefficient & 0.6\\
				$T$ & Projection window & 0.05 s\\
				$\tau$ & Blend time constant & 0.005 s\\
				$l$ & Pelvis target height & 0.85 m\\
				$T_{s}$ & Stance duration & 0.3 s\\
				$T_{f}$ & Flight duration & 0.09 s\\
				$\sigma_{s}$& Stance duration variance & 0.2\\
				$\sigma_{f}$& Flight duration variance & 0.1\\
				$c$ & Footstep lateral offset & 0.04 m\\
				$d$ & Foot clearance & 0.2 m\\
				$K_x$ & Raibert footstep sagittal feedback & 0.01\\
				$K_y$ & Raibert footstep lateral feedback & 0.3\\
				$W_u$ & Regularization weight on inputs & 1e-6\\
				$W_{acc}$ & Regularization weight on generalized accelerations $\ddot q$ & 1e-4\\
				$W_{\lambda}$ & Regularization weight on contact forces & 1e-1\\
				\midrule
			\end{tabular}
		}
		\label{tab:full_running_gains}
	\end{table}
	\begin{table}[h]
		\centering
		\caption{\change{Feedback gains for the Cassie running controller deployed on hardware. Weight and gain matrices are diagonal matrices, represented here as vectors to be concise.}}
		\resizebox{0.48\textwidth}{!}{
			\begin{tabular}{llll}
				\toprule
				\textbf{OSC Objective} & $W$ & $K_p$ & $K_d$ \\
				\midrule
				Toe joint angle &  0.01 & 1500 & 10\\
				Hip yaw angle & 2.5 & 100 & 5\\
				Pelvis [x, y, z] & $[0, 0, 5]$ & $[0, 0, 115]$ & $[0, 0, 5]$\\
				Pelvis [roll, pitch, yaw] & $[10, 5, 1]$ & $[150, 200, 0]$ & $[10, 10, 5]$\\
				Foot [x, y, z] & $[10, 100, 10]$ & $[125, 75, 75]$ & $[5, 5, 5]$\\
				\bottomrule
			\end{tabular}
		}
		\label{tab:running_gains}
	\end{table}
	
	\change{
		\subsection{Least Squares and Projections}
		\label{subsec:least_squares_and_projections}
		It is well studied that the least squares problem is equivalent to a projection \cite{strang1993introduction}.
		We show here that our choice of least squares in \cref{eq:ii_optimization_problem} is equivalent to the impact-invariant projection matrix in \cref{eq:ii_projection} when the tracking objectives are the generalized robot states, $y = q$, $\dot y = \dot q$, and there are no holonomic constraints $J_h = 0$.
		Recall, $J_y = \PartialDiff{y}{q}$, so in this case $J_y = \PartialDiff{q}{q} = I$, where $I$ is the identity matrix.
		We can simplify \cref{eq:kkt_closed_form} using these assumptions to arrive at
		\begin{align*}
			\begin{bmatrix}
				\lambda^*
			\end{bmatrix} = \begin{bmatrix}
				A^T A 
			\end{bmatrix}^{-1} \begin{bmatrix}
				A^T (\dot q_{des} - \dot q).
			\end{bmatrix} \label{eq:kkt_simplified},
		\end{align*}
		where $A = M^{-1} \contactJacobian^T$.
		We can substitute $\lambda^*$ into \cref{eq:vel_proj}
		\begin{align*}
			\dot q_{proj} = \dot q + A (A^T A)^{-1} A^T (\dot q_{des} - \dot q)
		\end{align*}
		to arrive at the projected generalized velocities.
		$\dot y_{proj} = \dot q_{proj}$ so we can express the projected velocity tracking error $\dot{\tilde{y}}_{proj}$ as:
		\begin{align*}
			\dot{\tilde{y}}_{proj} &= \dot q_{des} - (\dot q + A (A^T A)^{-1} A^T (\dot q_{des} - \dot q))\\
			&= (I -  A (A^T A)^{-1} A^T) (\dot q_{des} - \dot q)\\
			&= Q (\dot q_{des} - \dot q),
		\end{align*}
		which is identical to the result from the impact-invariant projection.
	}
	
	\change{
		\subsection{Modeling Considerations for Simulation}
		The running controller gains tuned in simulation are able to directly transferred to hardware; however, this is only possible after carefully modeling components such as reflected inertia and motor models to make the Cassie simulator in Drake better match the hardware.
		We use an approximation of the reflected inertia \cite{featherstone2014rigid}, which adds inertia contributions from the actuator rotor inertias directly to the corresponding diagonal terms of the mass matrix $M$.
		Instead of static actuator limits, we use a standard torque-speed curve to dynamically limit the torques provided by the actuators as a linear function of the actuator speed.
		We empirically observe that the running controller frequently operates at these dynamic actuator limits rather than the default static limits imposed by most simulators.
	}
	
	\subsection{Hardware Setup}
	\label{subsec:hardware_details}
	
	All processing is done on the Intel NUC 11 computer onboard Cassie. Note we swapped the original Intel NUC onboard Cassie to take advantage of the faster compute of a newer CPU.
	We run the state estimator and the controllers asynchronously as separate processes, and communication between processes is handled using LCM \cite{huang2010lcm}, while user commands are sent through the radio remote.
	
	\subsubsection{State Estimator}
	We use the contact-aided invariant EKF developed in \cite{hartley2020contact} to estimate the floating-base pelvis state.
	Although we do not directly use contact detection in our controllers, the state estimator utilizes the current contact mode estimate in the measurement update.
	We achieve this using a generalized-momentum observer, similar to the method used in \cite{bledt2018contact}, to estimate the contact force at each foot.
	We then set a threshold of 60 Nm on the contact normal force to define contact.
	We observe that this has a faster response and better accuracy over detecting contact using spring deflections.
	The state estimator runs at 2000 Hz.
	
	\subsubsection{Controller Implementation}
	We write our controllers using Drake \cite{drake} for the systems framework and all rigid body kinematics calculations.
	To set different tracking priorities during the flight phase for the running controller, we linearly ramp the weight for the foot tracking objective from 0.5 to 4 times the specified value across the duration of the trajectory.
	Although the tracking objectives are expressed in the world frame, we compute the errors in the robot yaw frame in order to have the gains operate on the sagittal and lateral directions rather than the world x and y directions.
	Finally, we solve the OSC QP using a minor modification of the OSQP \cite{stellato2020osqp} interface provided by \cite{drake} at 1500 Hz.

\end{document}